\PassOptionsToPackage{normalem}{ulem}
\PassOptionsToPackage{table,dvipsnames}{xcolor}
\documentclass[]{arxiv_template}

\usepackage[utf8]{inputenc}
\usepackage{times}
\usepackage{latexsym}
\usepackage{inconsolata}

\usepackage{graphicx}
\graphicspath{{figures/}}
\usepackage{amsmath}
\usepackage{amssymb}
\usepackage{amsfonts}
\usepackage{nicefrac}
\usepackage{booktabs}
\usepackage{multirow}
\usepackage{makecell}
\usepackage{enumitem}
\usepackage{longtable}
\usepackage{array}
\usepackage{float}
\usepackage{pifont}
\usepackage{xspace}
\newcolumntype{L}[1]{>{\raggedright\arraybackslash}p{#1}}
\tcbuselibrary{listings,breakable}

\newcommand{\safeincludegraphics}[2][]{%
  \IfFileExists{#2}{%
    \includegraphics[#1]{#2}%
  }{%
    \fbox{%
      \begin{minipage}[c][0.22\textheight][c]{0.9\linewidth}
        \centering
        \textbf{Missing figure}\\[0.5em]
        \texttt{\detokenize{#2}}
      \end{minipage}%
    }%
  }%
}

\newcommand{\placeholdertext}[1]{%
  \textcolor{red}{\textbf{[\texttt{\detokenize{#1}}]}}%
}
\newcommand{\placeholder}[1]{%
  \ifmmode
    \text{\placeholdertext{#1}}%
  \else
    \placeholdertext{#1}%
  \fi
}

\definecolor{headerblue}{RGB}{40, 80, 140}
\definecolor{headerbg}{RGB}{220, 235, 255}
\definecolor{closedtag}{RGB}{245, 240, 255}
\definecolor{opentag}{RGB}{240, 255, 240}
\definecolor{rowgray}{RGB}{248, 248, 252}
\definecolor{highlight}{RGB}{220, 190, 255}
\definecolor{best}{RGB}{0, 140, 100}
\definecolor{second}{RGB}{30, 100, 180}
\definecolor{genhead}{RGB}{52, 110, 183}
\definecolor{genbg}{RGB}{227, 238, 252}
\definecolor{edithead}{RGB}{46, 139, 87}
\definecolor{editbg}{RGB}{228, 248, 235}
\definecolor{repairhead}{RGB}{192, 72, 54}
\definecolor{repairbg}{RGB}{252, 232, 228}
\definecolor{rowalt}{RGB}{248, 249, 252}

\definecolor{goldbg}{RGB}{255,238,170}
\definecolor{silverbg}{RGB}{228,232,238}
\definecolor{bronzebg}{RGB}{246,222,200}
\newcommand{\goldc}[1]{\cellcolor{goldbg}\textbf{#1}}
\newcommand{\silverc}[1]{\cellcolor{silverbg}#1}
\newcommand{\bronzec}[1]{\cellcolor{bronzebg}#1}

\definecolor{promptbg}{RGB}{248,249,252}
\definecolor{promptframe}{RGB}{52,110,183}
\definecolor{promptsep}{RGB}{180,200,230}

\newtcblisting{promptbox}[1]{%
  listing only,
  colback=promptbg,
  colframe=promptframe,
  title={\small\bfseries #1},
  fonttitle=\small\bfseries,
  listing options={%
    basicstyle=\ttfamily\scriptsize,
    breaklines=true,
    breakautoindent=false,
    columns=fullflexible,
    keepspaces=true,
    aboveskip=0pt,
    belowskip=0pt,
  },
  boxrule=0.6pt,
  arc=2pt,
  left=6pt, right=6pt, top=2pt, bottom=2pt,
  breakable,
}

\tcbset{
  promptstyle/.style={
    listing only,
    colback=promptbg,
    colframe=promptframe,
    fonttitle=\small\bfseries,
    listing options={
      basicstyle=\ttfamily\scriptsize,
      breaklines=true,
      breakautoindent=false,
      columns=fullflexible,
      keepspaces=true,
      aboveskip=0pt,
      belowskip=0pt,
    },
    boxrule=0.6pt,
    arc=2pt,
    left=6pt, right=6pt, top=2pt, bottom=2pt,
    breakable,
  },
}

\fancypagestyle{titlestyle}{
  \fancyhf{}
  \fancyhead[L]{\small\sffamily\color{themecolor}\leftmark}
  \fancyhead[R]{\small\sffamily\color{themecolor}ANCHOR}
  \fancyfoot[C]{\small\sffamily\thepage}
  
  \renewcommand{\headrule}{
  \vskip -2mm
  \hbox to\headwidth{\color{themecolor}\leaders\hrule height 0.5pt\hfill}}
}

\title{ANCHOR: Assessing Naturalistic Constraint Hierarchies and Ordering in Repository-Grounded Agentic Coding}
\makeatletter
\renewcommand{\titlelist}{{\titlefont\rmfamily\bfseries\thetitle\par}}
\makeatother

\makeatletter
\gdef\authorlist{%
  \begin{center}
    \authorfont
    Wenqiang Zhu$^{1,*}$\quad
    Xinping Lei$^{2,*}$\quad
    Zhiyi Lai$^{3}$\quad
    Zhengshuo Gong$^{2}$\\[2pt]
    Tianzhuang He$^{2}$\quad
    Songwei Yu$^{4}$\quad
    Weihao Li$^{1}$\quad
    Jingxuan Xu$^{4}$\\[2pt]
    Zihang Xu$^{5}$\quad
    Ziqi Miao$^{3}$\quad
    Gang Wu$^{3}$\quad
    Zhongqiu Wang$^{3}$\quad
    Yuhui Bai$^{3}$\\[2pt]
    Jie Li$^{3}$\quad
    Zhao Liu$^{3}$\quad
    Junhao Fan$^{6}$\quad
    Yan Wu$^{3}$\\[2pt]
    Yanan Wu$^{2}$\quad
    Jiaheng Liu$^{2}$\quad
    Quan Gu$^{3}$\quad
    Ariel Deng$^{3,\dagger}$
  \end{center}
}
\makeatother

\makeatletter
\gdef\affiliationlist{%
  \begin{center}
    \affiliationfont
    $^{1}$Southern University of Science and Technology\quad
    $^{2}$Nanjing University\quad
    $^{3}$Independent Researcher\\
    $^{4}$National University of Singapore\quad
    $^{5}$Beihang University\quad
    $^{6}$University of Michigan
  \end{center}
}
\makeatother

\hypersetup{
  pdfauthor={Wenqiang Zhu, Xinping Lei, Zhiyi Lai, Zhengshuo Gong, Tianzhuang He, Songwei Yu, Weihao Li, Jingxuan Xu, Zihang Xu, Ziqi Miao, Gang Wu, Zhongqiu Wang, Yuhui Bai, Jie Li, Zhao Liu, Junhao Fan, Yan Wu, Yanan Wu, Jiaheng Liu, Quan Gu, Ariel Deng}
}

\abstract{For repository-grounded coding agents, a functionally correct patch is not
necessarily an acceptable one: developers also constrain \emph{how} the task
must be carried out---modification scope, workflow order, validation, tool
use, and reporting. This capability, which we call \emph{engineering
constraint following}, remains largely invisible to current evaluations.
We introduce \textbf{ANCHOR} (\textbf{A}ssessing \textbf{N}aturalistic
\textbf{C}onstraint \textbf{H}ierarchies and \textbf{O}rdering in
\textbf{R}epository-Grounded Agentic Coding), a benchmark built on
naturalistic, multilingual repository tasks spanning eight language
groups. ANCHOR combines (i) a hierarchical taxonomy of 10 categories and
78 leaf types distilled from real developer interactions, (ii) an automated
pipeline that instantiates task-grounded atomic constraints and composes
them into typed \texttt{AND}/\texttt{THEN}/\texttt{PRIORITY} relations, and
(iii) a trajectory-grounded evaluation framework that checks both the code
diff and the execution trajectory with rule-based and LLM judges.
Across 19 agent configurations spanning 2 scaffolds and 10 foundation
models on 358 tasks, the strongest measured system, Claude Opus 4.8 under
Cline, reaches 70.1\% atomic compliance and 63.7\% relation compliance,
while 19.0\% of tasks pass every scored checklist item under the strict
all-item FCCR criterion. A five-engineer audit of
constraints from 50 instances reports 94\% feasibility, 94\% repository
consistency, and 96\% realism. In the primary paired set (14 configurations),
the mean signed change in task completion after adding constraints is
$+0.7$ percentage points; individual changes vary in both directions.
These results distinguish functional success from constraint compliance,
with failures concentrated in workflow ordering, tool-use discipline, and
reporting, and substantial interactions between models and scaffolds.
}
\checkdata[Dataset]{\url{https://huggingface.co/datasets/Yone-01/ANCHOR}}
\checkdata[Code]{\url{https://github.com/bupterlxp/ANCHOR}}

\begin{document}
\maketitle

{
\renewcommand{\thefootnote}{\fnsymbol{footnote}}
  \footnotetext[1]{Equal contribution.}
  \footnotetext[2]{Corresponding author.}
  \footnotetext[0]{Contact: \email{12111009@mail.sustech.edu.cn}.}
}

\section{Introduction}
\label{sec:intro}

Software engineering tasks are rarely unconstrained objectives. Developers ask
agents not only to fix the bug, but also to preserve APIs, reproduce before
patching, run specific tests, avoid prohibited tools, and report what was
actually executed. For coding agents, success is therefore not merely to solve
the task, but to solve it under explicit engineering constraints.

\begin{figure*}[t]
\centering
\safeincludegraphics[width=0.90\textwidth]{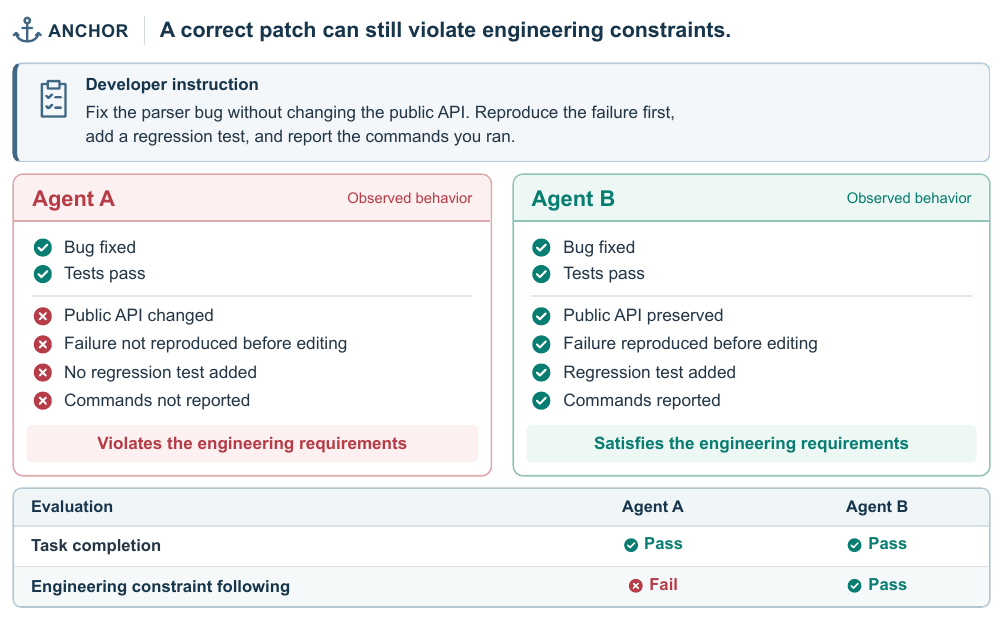}
\caption{\textbf{Illustrative schematic. A correct patch is not necessarily an acceptable engineering solution.}
Two agents fix the same bug and pass tests. Their observed behavior differs on API preservation, reproduction order, regression testing, and command reporting. ANCHOR distinguishes task completion from engineering constraint following.}
\label{fig:intro}
\end{figure*}

We argue that this is not simply generic instruction following with code
context. In repository-grounded agentic coding, many requirements are
trajectory-grounded: they constrain tool use, workflow order, validation,
and reporting over the course of execution rather than only the final artifact.
They are also often structured, because multiple requirements must be
satisfied jointly, in order, or under explicit priority. Engineering constraint
following is thus a distinct capability dimension, complementary to both task
completion and scaffold-level policy compliance. Figure~\ref{fig:intro}
illustrates this distinction: two trajectories may solve the same issue, yet
only one is acceptable under the stated engineering requirements.

This gap has become important because modern coding agents can already solve a
substantial fraction of real repository issues on SWE-bench-style
benchmarks~\citep{jimenez2024swebench,zan2025multiswebench,swecompass2024,swebenchv22025}.
Yet these benchmarks largely evaluate whether the final artifact is correct,
not whether the agent obeyed the task-level requirements that governed its
execution. General instruction-following benchmarks
do not capture repository-grounded, multi-step engineering workflows, while
recent agentic instruction-following benchmarks still emphasize generic agentic
instructions or scaffold-level policies rather than task-level
engineering requirements~\citep{zhou2023ifeval,jiang2024followbench,pyatkin2025ifbench,wu2025lifbench,zou2025eifbench,zhang2025cfbench,qi2025agentif,yao2024taubench,ding2026octobench}. Current evaluations may therefore overestimate real-world readiness.
Table~\ref{tab:benchmark_comparison} situates ANCHOR within this landscape.

\begin{table*}[t]
\centering
\caption{Comparison with prior instruction-following and coding benchmarks.}
\label{tab:benchmark_comparison}
{\small
\setlength{\tabcolsep}{4pt}
\renewcommand{\arraystretch}{1.16}
\resizebox{\textwidth}{!}{%
\begin{tabular}{lccccccccc}
\toprule
\rowcolor{headerbg}
\textbf{Benchmark} & \textbf{Task} & \textbf{Real} & \textbf{Scope} & \textbf{Comp.} & \textbf{Auto.} & \textbf{Traj.} & \textbf{Eval.} & \textbf{Scale} & \textbf{Lang.} \\
\midrule
IFEval & NL gen. & \textcolor{red!70}{\ding{55}} & \textcolor{red!70}{\ding{55}} & \textcolor{red!70}{\ding{55}} & \textcolor{red!70}{\ding{55}} & \textcolor{red!70}{\ding{55}} & Rule & 541 & EN \\
\rowcolor{rowgray}
FollowBench & NL gen. & \textcolor{red!70}{\ding{55}} & \textcolor{red!70}{\ding{55}} & \textcolor{red!70}{\ding{55}} & \textcolor{red!70}{\ding{55}} & \textcolor{red!70}{\ding{55}} & LLM & 820 & EN \\
ComplexBench & NL gen. & \textcolor{red!70}{\ding{55}} & \textcolor{red!70}{\ding{55}} & \textcolor{teal}{\ding{51}} & \textcolor{teal}{\ding{51}} & \textcolor{red!70}{\ding{55}} & LLM+Rule & $\sim$1K & ZH/EN \\
\rowcolor{rowgray}
BigCodeBench & Code gen. & \textcolor{red!70}{\ding{55}} & Limited & \textcolor{red!70}{\ding{55}} & \textcolor{red!70}{\ding{55}} & \textcolor{red!70}{\ding{55}} & Exec. & 1140 & Python \\
CIFE & Code gen. & \textcolor{red!70}{\ding{55}} & Artifact & \textcolor{red!70}{\ding{55}} & \textcolor{teal}{\ding{51}} & \textcolor{red!70}{\ding{55}} & LLM & 1000 & Python \\
\rowcolor{rowgray}
CodeIF & Code gen. & \textcolor{red!70}{\ding{55}} & Limited & \textcolor{red!70}{\ding{55}} & \textcolor{red!70}{\ding{55}} & \textcolor{red!70}{\ding{55}} & Rule & -- & Multi \\
AgentIF & Agentic & \textcolor{red!70}{\ding{55}} & Limited & \textcolor{red!70}{\ding{55}} & \textcolor{teal}{\ding{51}} & \textcolor{red!70}{\ding{55}} & LLM & 707 & EN \\
\rowcolor{rowgray}
OctoBench & Agentic & \textcolor{teal}{\ding{51}} & Limited & \textcolor{red!70}{\ding{55}} & \textcolor{red!70}{\ding{55}} & Limited & Rule & 217 & Multi \\
\midrule
\rowcolor{highlight}
\textbf{ANCHOR (Ours)} & \textbf{Agentic} & \textcolor{teal}{\ding{51}} & \textbf{Eng.} & \textcolor{teal}{\ding{51}} & \textcolor{teal}{\ding{51}} & \textcolor{teal}{\ding{51}} & \textbf{Rule+LLM} & \textbf{358} & \textbf{Multi} \\
\bottomrule
\end{tabular}
}
}
\end{table*}

To address this gap, we introduce \textbf{ANCHOR}
(\textbf{A}ssessing \textbf{N}aturalistic \textbf{C}onstraint
\textbf{H}ierarchies and \textbf{O}rdering in \textbf{R}epository-Grounded
Agentic Coding), a benchmark for evaluating engineering constraint following in
repository-grounded coding tasks. ANCHOR combines a 10-category, 78-leaf
taxonomy, an automated pipeline that instantiates atomic constraints and
composes them into structured \textsc{and}/\textsc{then}/\textsc{priority}
relations, and a trajectory-grounded evaluation framework with rule-based and
LLM judges. Across 19 constrained configurations, including a full-benchmark
Claude Opus 4.8 evaluation under Cline, the highest strict all-item FCCR is 19.0\%.
The primary paired set (14 configurations) shows a mean signed completion change
of $+0.7$ percentage points after adding constraints, with gains and losses
across systems. This calibration does not establish joint satisfiability or
equal task difficulty; the direct evidence is that functional completion and
checklist compliance differ within each evaluated configuration.
Our contributions are to define engineering constraint following,
operationalize it in a benchmark, and characterize the resulting failures,
supported by human audits of both generated constraints and judge decisions.

Benchmark instances are available on
\href{https://huggingface.co/datasets/Yone-01/ANCHOR}{Hugging Face}; the
construction/evaluation prompts and taxonomy are available in the
\href{https://github.com/bupterlxp/ANCHOR}{public repository}.

\section{Related Work}
\label{app:related}

Repository-grounded coding benchmarks such as SWE-bench, SWE-Compass, and
SWE-Rebench~V2 measure whether an agent can resolve real software engineering
tasks in real repositories, typically through final-patch correctness under an
executable test harness~\citep{jimenez2024swebench,swecompass2024,
swebenchv22025,zan2025multiswebench,deng2025swebenchpro,thai2026sweevo,yu2026swenexus,
featurebench2026,li2024devbench}. These benchmarks have driven rapid progress
in coding agents, but they primarily measure \emph{task completion}. They
therefore provide limited visibility into whether an agent respected
task-specific engineering requirements on scope, workflow, validation, or
reporting during execution. Complementary evaluations assess repository-level
completion~\citep{yang2024execrepobench,liu2025m2rc}, visual and interactive
artifact quality~\citep{lei2026webcompass,zhang2025artifactsbench}, and code
critique~\citep{zhang2025codecriticbench}. Trajectory diagnosis and process
feedback further expose intermediate agent failures~\citep{li2026codetracer,xu2026flare}.

Constraint-aware instruction-following benchmarks establish the importance of
explicitly verifiable requirements, both in natural language and in code
generation~\citep{zhou2023ifeval,jiang2024followbench,pyatkin2025ifbench,
wu2025lifbench,wen2024complexbench,chen2024sifo,qin-etal-2024-infobench,
zhang2025iheval,zhuo2024bigcodebench,yan2025codeif,gunnu2025cife}. However,
they mostly evaluate single-turn settings, final artifacts, or flat constraint
sets. Recent work such as Inverse IFEval further shows that even strong LLMs
can fail to override entrenched training conventions in order to follow explicit
user instructions~\citep{zhang2025inverse}. ANCHOR instead targets
repository-grounded agentic coding, where many requirements are trajectory-level
and many must be interpreted jointly through ordering or priority relations.

The closest concurrent work is OctoBench~\citep{ding2026octobench}, which
studies scaffold-aware instruction following in repository-grounded coding.
We view the two benchmarks as complementary rather than competing. OctoBench
targets \emph{scaffold- and platform-level policies}, whereas ANCHOR targets
\emph{task-level engineering constraints} specified by the developer
instruction itself. ANCHOR further contributes an explicit engineering
constraint taxonomy and structured composite relations, enabling direct
evaluation of hierarchy, ordering, and precedence.
Table~\ref{tab:benchmark_comparison} summarises this
positioning against prior instruction-following and coding benchmarks.

\section{Constraint Taxonomy}
\label{sec:taxonomy}

The ANCHOR taxonomy is constructed bottom-up from real developer interactions
with coding agents rather than from an a priori schema. We first collected a
corpus of agentic coding interactions---task instructions, inline review
comments, and post-hoc feedback on agent trajectories---and performed
grounded open-label annotation to extract the engineering constraints that
developers impose on coding agents in practice. A panel of senior software
engineers then iteratively refined this raw label space by clustering
near-duplicate labels, resolving boundary cases, and stress-testing the
emerging hierarchy against held-out trajectories. The resulting hierarchy
contains \textbf{10 top-level categories}, \textbf{34 mid-level
subcategories}, and \textbf{78 leaf categories}; in the current release, these
leaves account for \textbf{2,836} instantiated atomic constraints. Full
annotation guidelines, boundary cases, and the complete taxonomy are provided
in Appendix~\ref{app:taxonomy} and Appendix~\ref{app:construction_details}.

Throughout this refinement, the taxonomy is treated as an evaluation substrate
rather than a descriptive catalog, guided by five design principles.
\emph{Coverage}: the taxonomy should span the major constraint types observed
in real developer data, so that practical instructions map into it with
minimal residue. \emph{Semantic exclusivity}: leaf categories should be as
disjoint as possible, reducing ambiguity during annotation and automatic
mapping. \emph{Operationalisability}: each leaf should admit concrete,
testable instantiations with stable pass conditions under either rule-based or
LLM-based judging. \emph{Hierarchical organisation}: the structure should
support both fine-grained diagnosis at the leaf level and coarse-grained
reporting at the category level. \emph{Language neutrality}: the same
conceptual categories should transfer across the multilingual repository
settings covered by ANCHOR.

\begin{figure}[tbp]
\centering
\begin{minipage}[b]{0.485\textwidth}
    \centering
    \safeincludegraphics[width=\linewidth]{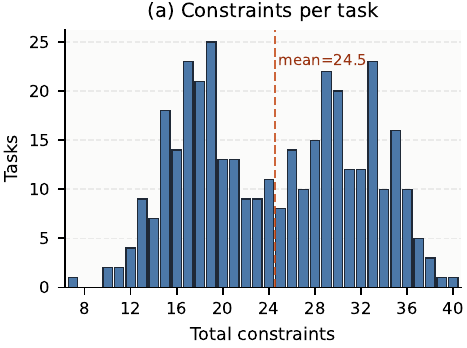}
\end{minipage}\hfill
\begin{minipage}[b]{0.485\textwidth}
    \centering
    \safeincludegraphics[width=\linewidth]{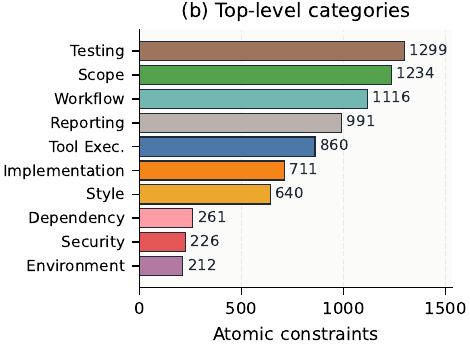}
\end{minipage}
\caption{Constraint footprint in the source developer queries. Left:
constraints per original query. Right: extracted atomic constraints across
the ten top-level categories. These panels describe the source-query
corpus; the instantiated benchmark taxonomy is shown separately in
Figure~\ref{fig:instantiated_taxonomy}.}
\label{fig:query_overview}
\end{figure}

\begin{figure}[tbp]
\centering
\safeincludegraphics[width=0.98\linewidth]{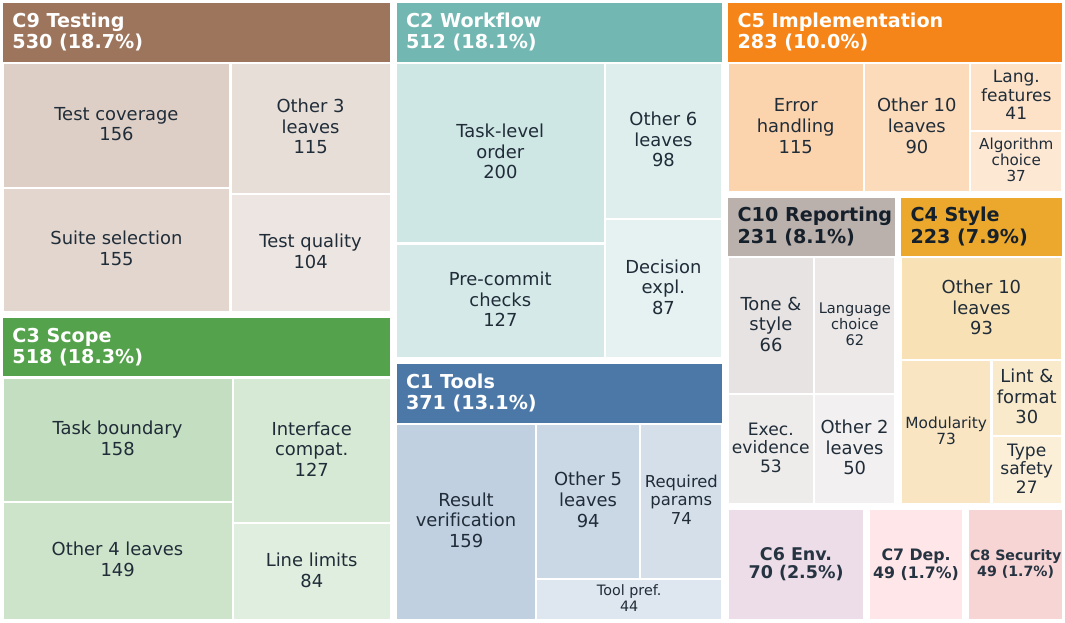}
\caption{Instantiated ANCHOR taxonomy: 2,836 atomic constraints across ten
categories. Category areas are proportional to their atomic counts. The
three largest leaf types are labelled within each large category; the
remaining leaves are combined into an explicitly labelled ``Other'' tile.
The three smallest categories show their totals. Colours match
Figure~\ref{fig:query_overview}; Table~\ref{tab:taxonomy-leaf} lists all 78 leaves.}
\label{fig:instantiated_taxonomy}
\end{figure}

\begin{table}[tbp]
  \centering
  \caption{Top-level engineering constraint taxonomy in ANCHOR, with leaf counts and instantiated atomic-constraint counts from the current release.}
  \label{tab:taxonomy-top}
  \small
  \setlength{\tabcolsep}{4pt}
  \begin{tabular}{clcc}
    \toprule
    \rowcolor{headerbg}
    \textbf{Code} & \textbf{Category} & \textbf{\#Leaves} & \textbf{\#Atomic} \\
    \midrule
    C1 & Tool Execution Protocol             & 8  & 371 \\
    C2 & Workflow \& Behaviour               & 9  & 512 \\
    C3 & Modification Scope \& Change Budget & 7  & 518 \\
    C4 & Code Style \& Engineering Standards & 13 & 223 \\
    C5 & Code Implementation                 & 13 & 283 \\
    C6 & Environment                         & 6  & 70  \\
    C7 & Dependency \& Build Strategy        & 5  & 49  \\
    C8 & Security \& Dangerous Operations    & 6  & 49  \\
    C9 & Testing                             & 6  & 530 \\
    C10 & Output Format \& Reporting         & 5  & 231 \\
    \midrule
    \textbf{Total} &                         & \textbf{78} & \textbf{2,836} \\
    \bottomrule
  \end{tabular}
\end{table}

At the top level, the ten categories capture complementary dimensions of
agentic coding. \textbf{C1 (Tool Execution Protocol)} governs how tools are
invoked, including tool selection, invocation parameters, failure handling,
and result verification, and \textbf{C2 (Workflow \& Behaviour)} governs how
the agent conducts the task over time, including execution order, user
confirmation, information disclosure, and commit discipline. \textbf{C3
(Modification Scope \& Change Budget)} bounds the spatial, quantitative, and
semantic extent of changes, \textbf{C4 (Code Style \& Engineering Standards)}
covers formatting, naming, type-system usage, and documentation, and
\textbf{C5 (Code Implementation)} constrains implementation choices such as
algorithms, error handling, side-effect control, and technology selection.
\textbf{C6--C8} specify environment compatibility and isolation, dependency
and build policy, and prohibitions on dangerous or privileged operations,
while \textbf{C9 (Testing)} and \textbf{C10 (Output Format \& Reporting)}
impose validation and reporting requirements. Leaf categories are the smallest
operational unit: each captures a semantically coherent developer requirement
that can be instantiated into task-grounded constraints and verified with
relatively stable logic. The categories are complementary rather than
alternative---a single instruction such as ``only modify files under
\texttt{common/crypto/}, reproduce the bug before applying the fix, and add a
regression test before reporting completion'' simultaneously invokes C3
(scope), C2 (workflow), and C9 (testing).

Figure~\ref{fig:query_overview} summarizes the source-query corpus, while
Figure~\ref{fig:instantiated_taxonomy} and Table~\ref{tab:taxonomy-top}
describe the instantiated benchmark. As a lightweight sanity check against template collapse, we mask
repository names, file paths, function names, and other task-specific
identifiers, then measure pairwise similarity among constraints instantiated
from the same leaf. Across 130,049 within-leaf pairs drawn from 70
multi-instance leaves, the mean masked cosine similarity is 0.565, indicating
shared semantics within a leaf without near-duplicate templating.

\section{ANCHOR Construction Pipeline}
\label{sec:pipeline}

\begin{figure}[t]
    \centering
    \safeincludegraphics[width=\linewidth]{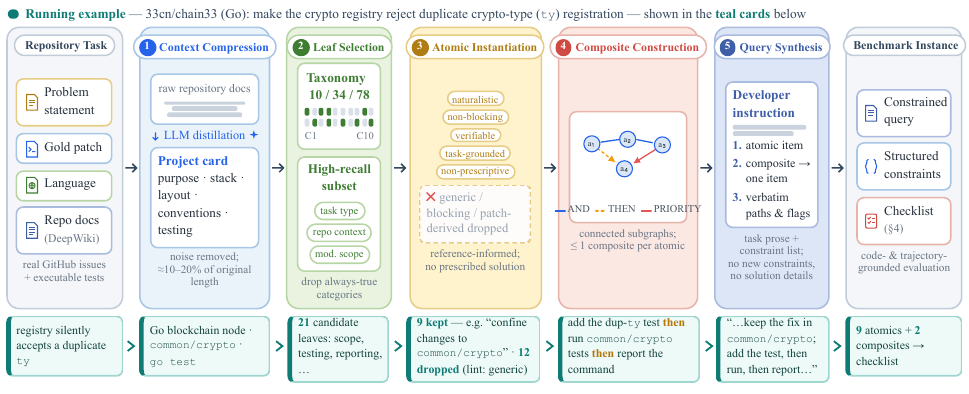}
    \caption{Overview of the ANCHOR construction pipeline. Starting from a
    repository task drawn from real GitHub issues (problem statement, gold
    patch, language, repository documentation), five stages produce each
    benchmark instance: (1) repository documentation is distilled into a
    compact project card; (2) a high-recall subset of taxonomy leaves is
    selected from task, repository, and scope evidence; (3) leaves are
    instantiated into task-grounded atomic constraints under explicit
    quality gates; (4) related atomics are composed into connected
    composite graphs with typed \texttt{AND}/\texttt{THEN}/\texttt{PRIORITY}
    relations; and (5) the result is rendered as a natural developer
    instruction paired with a structured constraint object, which
    Section~\ref{sec:evaluation_framework} operationalizes into an
    executable checklist. A running example from \texttt{33cn/chain33}
    (rejecting duplicate crypto-type registration in Go) is threaded
    through the figure as the teal example lane at the bottom.}
    \label{fig:pipeline}
\end{figure}

ANCHOR draws repository-level coding environments from the multilingual
benchmark SWE-Compass~\citep{swecompass2024}, which is built from real GitHub
issues and provides gold-standard patches with executable test environments
across Python, Go, Java, TypeScript, Rust, C, C++, and C\#. We retain
instances that satisfy four properties: (i) a usable execution environment,
enabling trajectory-grounded evaluation in a unified sandbox; (ii) sufficient
task and repository context to support downstream constraint selection; (iii)
moderate patch complexity, excluding both trivial micro-edits and
long-horizon changes that are difficult to evaluate consistently; and (iv)
broad coverage across programming languages and repository types. The purpose
of this filtering is not to favour easier tasks, but to ensure that each
retained instance can support realistic, task-grounded, and verifiable
engineering constraints.
Figure~\ref{fig:pipeline} summarizes how ANCHOR converts a raw repository task
into both a natural developer instruction and a structured constraint object
that can be evaluated after agent execution, through five explicit steps.

\paragraph{Step 1: Context collection and compression.}
For each retained task, we gather the repository identifier, target language,
original problem statement, gold patch, and repository-level documentation
from DeepWiki~\citep{deepwiki2025}. Raw documentation is often too long and
noisy to include directly in downstream prompts, so we compress it into a
compact project-context card that preserves repository purpose, technology
stack, directory structure, engineering conventions, testing practices, and
version information, while discarding diagrams, verbose API documentation,
illustrative code examples, and marketing language. This is a targeted
information-distillation step rather than mere length reduction: the card
supplies repository-level evidence for later decisions, such as whether
linting, testing, or dependency-policy constraints are meaningful for a given
project.

\paragraph{Step 2: Leaf-category selection.}
Given the 78 leaf categories, this stage retrieves a high-recall subset of
potentially relevant leaves for the task, based on three complementary
evidence sources: \emph{task-type evidence} (e.g., testing and
interface-compatibility constraints are natural in bug-fixing tasks),
\emph{repository-context evidence} (e.g., explicit formatter, linter, or
type-checking conventions make the corresponding categories meaningful), and
\emph{modification-scope evidence} (e.g., tasks touching public interfaces,
filesystem logic, or security-sensitive code raise different constraint
families). The selector deliberately avoids categories that are only
theoretically applicable to almost any coding task, since these tend to yield
generic and weakly discriminative constraints later on.

\paragraph{Step 3: Atomic constraint instantiation and filtering.}
Each selected leaf is instantiated into zero or more \emph{atomic
constraints}---concrete, task-specific constraint statements grounded in the
actual files, tools, and technologies involved. The selection and
instantiation prompts receive the gold patch as context, including affected
modules and plausible implementation concerns. They explicitly prohibit
copying its exact file/line counts or restating its implementation as a
constraint; an engineering requirement should admit valid implementations
beyond the reference patch. Because instantiation and composition share a
prompt, the patch is also visible at composition time. These safeguards
reduce, but do not rule out, indirect solution leakage. Every retained constraint must be
\emph{naturalistic} (plausibly stated by a senior engineer assigning the
task), \emph{non-blocking} (constraining how the task is done without making
it infeasible), \emph{specific and verifiable} (admitting an objective
pass/fail judgement), \emph{task-grounded} (referencing concrete task
details), and independent in content from the gold solution: the generator
may inspect the patch for task context, but the final constraint should not
encode patch-specific file/line counts or implementation details.
The generator is instructed to discard candidates that fail these
criteria, so Steps 2 and 3 jointly form a recall--precision
cascade: coarse selection avoids prematurely missing meaningful categories,
while instantiation enforces quality control over what enters the benchmark.

\paragraph{Step 4: Composite construction.}
Real developer instructions are often not flat lists of independent
requirements; they specify joint satisfaction, required execution order, or
precedence among competing requirements. ANCHOR captures this structure by
composing task-grounded atomics into connected composite graphs whose edges
carry one of three typed relations: \texttt{AND} requires joint satisfaction;
\texttt{THEN} requires both satisfaction and correct temporal
order---reaching the right end state in the wrong order is scored as a
failure; and \texttt{PRIORITY} places two constraints intentionally in
tension, and a \texttt{PRIORITY} edge passes if and only if the
higher-priority source constraint is upheld, regardless of whether the
lower-priority target is sacrificed to it. This edge-level rule does not
waive other requirements in a mixed-relation composite. Edge types may be mixed within a single composite, so
multi-step and prioritised workflows can be expressed in one graph.
Composites are introduced only when the underlying relations are
semantically natural and task-meaningful, and atomics that do not admit such
structure remain standalone instruction items. Each atomic constraint
participates in at most one composite, and the connected-subgraph
formulation keeps each composite interpretable as a single instruction unit
rather than an arbitrary bundle of clauses.

The non-blocking criterion is a construction safeguard for individual
constraints, not a formal proof that all constraints can be jointly
satisfied. \texttt{AND} and \texttt{THEN} require compatible requirements;
\texttt{PRIORITY} intentionally introduces a conflict together with an
explicit precedence rule. Assigning an atomic constraint to at most one
composite limits graph overlap, but does not eliminate semantic interactions
between distinct composites. Small completion-rate changes likewise cannot
establish satisfiability, since an agent can solve the task while ignoring a
constraint. We therefore complement these safeguards with a separate human
audit: five senior engineers independently assessed the atomic constraints
in 50 instances for feasibility, repository consistency, and realism, with
reported validity rates of 94\%, 94\%, and 96\%, respectively.
Section~\ref{app:constraint_audit} provides the rubric and agreement
statistics; these results characterize sampled constraint quality rather
than certify every full constraint set.

\paragraph{Step 5: Query synthesis.}
The final stage renders the structured representation as a natural developer
instruction in two parts: a short task description rewritten in the voice of
a developer assigning work, followed by a constraint list of separable
instruction items. Independent atomics become individual items, while each
composite is rendered as a single natural-language item that preserves the
relation semantics among its members. Synthesis must preserve specificity:
exact file paths, command names, flags, module references, and behavioural
conditions from the instantiated constraints remain intact. It must also
neither introduce new constraints nor leak gold-solution details---the query
expresses requirements, not answers.

This five-stage design preserves the natural surface form of developer
instructions while grounding evaluation in an explicit, machine-checkable
representation of what the agent was required to do and in what relational
structure. Detailed filtering criteria, design rationales, prompt templates,
and stage-specific schemas are provided in
Appendix~\ref{app:construction_details} and
Appendix~\ref{app:prompts:compress}--\ref{app:prompts:query}.
Section~\ref{sec:evaluation_framework} then operationalizes this structured
representation into executable checklist items for code- and
trajectory-grounded evaluation.

\subsection{Human Audit of Generated Constraints}
\label{app:constraint_audit}
This audit examines the generated requirements themselves, separately from
the 480-item judge--human study in Section~\ref{app:judge_reliability}.
We sampled 50 benchmark instances. All five engineers independently
labelled every atomic constraint in these instances on three binary
dimensions, without seeing one another's labels. They were the same senior
engineers used in the judge study, with 6--12 years of experience: two
Python/Go backend engineers, one infrastructure engineer with Java/Bash
and CI/CD expertise, one TypeScript/Rust full-stack engineer, and one
research engineer with broad open-source code-review experience.

The rubric was as follows. \textbf{Feasibility}: PASS if the requirement
can be met in the given repository and task without preventing completion;
FAIL for an unsatisfiable or task-blocking requirement, such as restricting
edits to a directory that excludes the relevant files.
\textbf{Repository consistency}: PASS if referenced paths, symbols,
tools, dependencies, and conventions agree with the repository; FAIL for
hallucinated referents or an inapplicable toolchain.
\textbf{Realism}: PASS if a senior engineer could plausibly impose the
requirement here; FAIL if it is artificial, trivially satisfied, or merely
restates the task or reference solution.

Table~\ref{tab:constraint_audit} summarizes the five-annotator audit.
Raw agreement ranges from 92.0\% to 97.6\%, and Gwet's AC1 ranges from
0.91 to 0.97. These results support consistency of the sampled quality
judgments, while the audit covers atomic constraints from 50 instances
rather than every composite or complete task-level constraint set.
The validity rates do not estimate how much FCCR would change after
repairing ambiguous items.

\begin{table}[htbp]
\centering
\caption{Audit of atomic constraints from 50 instances, independently
labelled by five engineers. Validity and raw agreement are percentages; Gwet
AC1 is a coefficient.}
\label{tab:constraint_audit}
\small
\begin{tabular}{lrrr}
\toprule
Dimension & Validity & Raw agreement & Gwet AC1 \\
\midrule
Feasibility & 94.0 & 92.0 & 0.91 \\
Repository consistency & 94.0 & 96.4 & 0.96 \\
Realism & 96.0 & 97.6 & 0.97 \\
\bottomrule
\end{tabular}
\end{table}

\section{Evaluation Framework}
\label{sec:evaluation}
\label{sec:evaluation_framework}

\begin{figure}[t]
\centering
\safeincludegraphics[width=\linewidth]{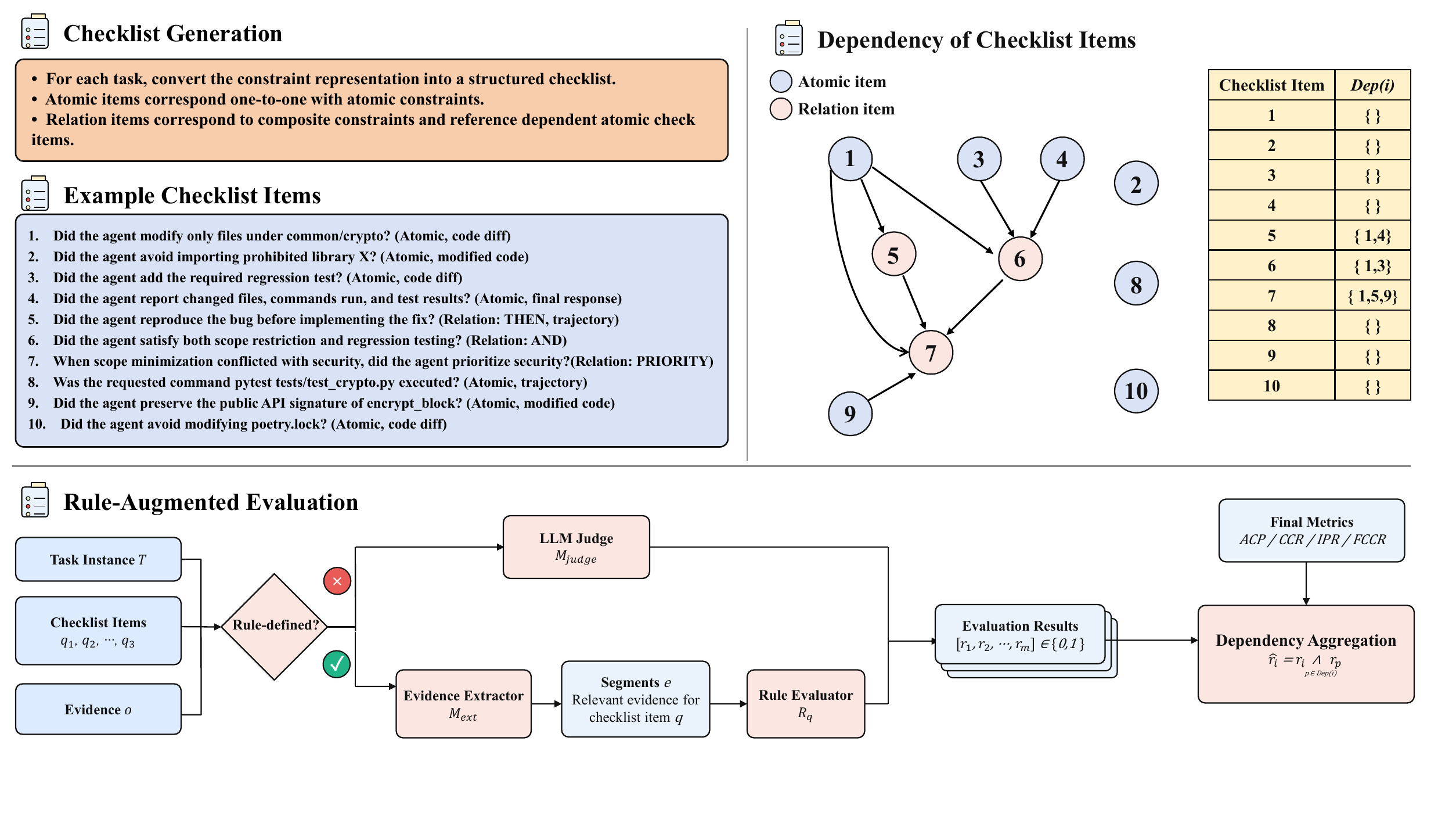}
\caption{Overview of the ANCHOR evaluation framework.}
\label{fig:evaluation_pipeline}
\end{figure}

Section~\ref{sec:pipeline} produces, for each task, a natural developer
instruction paired with a structured constraint representation. We now
describe how that representation is operationalized into executable checks
over the agent's code diff and trajectory.
Figure~\ref{fig:evaluation_pipeline} summarizes this process.

\paragraph{Checklist generation.}
For each task instance, we generate a \textbf{checklist}: a structured set of
evaluation items generated from the constraint representation and checked
against the final developer instruction, which is authoritative if the two
differ. This intermediate checklist layer is necessary because the
construction-time representation is convenient for instantiation and
composition, but evaluation requires item-level pass conditions, judge
assignments, and explicit evidence targets. Only constraints preserved in
the final instruction are retained. Verified atomic constraints yield
\emph{atomic items} one-to-one, including atomics that participate in
composites; each composite constraint additionally yields a \emph{relation
item} that references its member atomic items as dependent checks and
evaluates the higher-order property---joint satisfaction, temporal
ordering, or priority handling---on top of them. Each item contains a natural-language question, a pass
condition, a judge type with its parameters, and an evidence type specifying
whether the item should be checked against the code diff, the execution
trajectory (including the final response), or commit messages. This design supports
fine-grained diagnosis of individual violations as well as explicit
evaluation of structured constraint compliance.

\paragraph{Hybrid judging.}
When the pass condition is reducible to symbolic checks, ANCHOR uses
rule-based judges over explicit evidence sources; Table~\ref{tab:judges}
summarizes the rule templates. A single item may bind several rule checks
that must jointly hold---for example, verifying that one module's import is
absent while its replacement's import is present. Otherwise---especially for semantically
complex atomic items and relation items involving \texttt{THEN} or
\texttt{PRIORITY}---ANCHOR uses an LLM judge, which receives the checklist
question, the pass condition, and a task-relevant evidence excerpt, and
returns a binary \texttt{PASS}/\texttt{FAIL} decision with a brief
justification. Evidence excerpts are used instead of full raw trajectories
whenever possible to reduce noise and context length. This hybrid split
matters because many engineering constraints are only partially observable
from the final patch: scope, file-budget, and import restrictions are often
reducible to symbolic checks, whereas workflow order, reporting fidelity, and
relations such as \texttt{THEN} or \texttt{PRIORITY} require reasoning over
temporally ordered evidence extracted from the trajectory. Prompt templates
are provided in Appendix~\ref{app:prompts:checklist} and
Appendix~\ref{app:prompts:judge}.

\begin{table}[t]
  \centering
  \small
  \caption{Rule-based judge function types used in ANCHOR.}
  \label{tab:judges}
  \begin{tabular}{ll}
    \toprule
    \rowcolor{headerbg}
    \textbf{Judge type} & \textbf{Description} \\
    \midrule
    \texttt{path\_filter}      & Whether modified paths satisfy a whitelist/blacklist \\
    \texttt{file\_count}       & Whether the number of modified files satisfies a threshold \\
    \texttt{line\_count}       & Whether the number of changed lines satisfies a threshold \\
    \texttt{command\_presence} & Whether a command appears or does not appear in the trajectory \\
    \texttt{command\_arg\_check} & Whether a command uses required or forbidden flags \\
    \texttt{action\_sequence} & Whether observable actions occur in the required order \\
    \texttt{import\_check}     & Whether a module is imported or explicitly not imported \\
    \texttt{regex\_match}      & Whether target text matches a regular expression \\
    \bottomrule
  \end{tabular}
\end{table}

\subsection{Judge Validation}
\label{app:judge_reliability}
To assess the reliability of the LLM judge used throughout ANCHOR, we conducted a human agreement study on a stratified sample of 480 checklist instances drawn from held-out tasks, balanced across the ten top-level taxonomy categories, the three relation types (\textsc{And}, \textsc{Then}, \textsc{Priority}), and three difficulty strata (rule-verifiable, semantically straightforward, and semantically complex). Each (constraint, evidence) pair was independently labelled \textsc{Pass}/\textsc{Fail} by five human evaluators who did not participate in benchmark construction, taxonomy design, or prompt authoring. The evaluators were senior software engineers with 6--12 years of professional development experience (mean 8.4 years), drawn from a mix of backgrounds to reduce single-stack bias: two backend engineers with primary experience in Python and Go service development, one infrastructure engineer with CI/CD and build-system expertise across Java and Bash toolchains, one full-stack engineer working primarily in TypeScript and Rust, and one researcher--engineer with prior experience reviewing open-source pull requests at scale across multiple language ecosystems. All evaluators had prior code-review responsibility on production codebases of at least 100k lines, ensuring familiarity with the kinds of engineering constraints---modification scope, validation discipline, dependency hygiene, reporting obligations---that ANCHOR targets. Evaluators were given the checklist question, the explicit pass condition, and the same evidence excerpt provided to the LLM judge, but no access to the judge's output, in order to avoid anchoring. The majority label across the five evaluators was used as the human consensus, and inter-evaluator agreement on the sample itself was $\kappa = 0.83$, indicating that the consensus label is itself stable.

Aggregating across all sampled items, the LLM judge reaches Cohen's $\kappa = 0.78$ with the human consensus label, corresponding to $89.4\%$ raw agreement. Agreement is highest on rule-verifiable atomic items in categories such as Modification Scope (C3), Dependency (C7), and Tool Execution Protocol (C1), with $\kappa = 0.86$, where the pass condition reduces to a deterministic check over the diff or trajectory. Agreement remains substantial on semantically straightforward items at $\kappa = 0.79$, and is lowest, though still acceptable, on relation items involving \textsc{Then} ordering and \textsc{Priority} conflict resolution at $\kappa = 0.69$. The lower agreement on these relations is relevant when interpreting
Section~\ref{sec:experiments:composite}: ordering is difficult both to follow
and to verify, and the present alignment study does not fully separate
agent errors from evaluation noise.

Disagreement is concentrated in three patterns rather than spread uniformly across the sample. First, \textbf{partial-evidence cases} account for $41.2\%$ of disagreements: the LLM judge tends to mark an item \textsc{Pass} when some---but not all---sub-components of the pass condition are demonstrated in the evidence, whereas human evaluators apply a stricter all-or-nothing reading. This is most common in C9 (Testing) and C10 (Reporting), where a constraint such as \emph{``report changed files, commands run, and test results''} is partially fulfilled by the agent's final summary. Second, \textbf{temporal-boundary cases} account for $26.7\%$: for \textsc{Then} relations, the judge occasionally accepts an interleaved or out-of-order trajectory as compliant if both component actions are eventually present, whereas human evaluators enforce strict precedence (e.g., reproduction must occur before any modification, not merely before the final patch). Third, \textbf{intent-versus-action cases} account for $19.4\%$: when the agent verbally acknowledges a constraint in its final report but does not in fact act on it, the LLM judge is somewhat more lenient than human evaluators, who consistently mark such cases \textsc{Fail}. The remaining $12.7\%$ of disagreements are scattered across category-specific edge cases without a dominant pattern.

In a human-subset consistency check, the configurations ranked on the
480-item human-labelled subset have Spearman correlation $0.96$ with the
full judge-based ranking. This comparison also changes
the evaluated item set and is not an isolated judge-swap experiment. No
cross-provider judge comparison or controlled prompt-sensitivity study is
reported. Consequently, the alignment results do not establish absence of
model-family bias or robustness to arbitrary judge/prompt choices. We treat
the LLM judge as an imperfect proxy and report its lower agreement on hard
relations explicitly.

\subsection{Metrics}
\label{sec:eval:metrics}

We report four complementary metrics. For a benchmark set with $N$ task
instances, let $A_t$ and $R_t$ denote the numbers of atomic and relation items
for task $t$, let $\mathbb{I}[\cdot]$ be the indicator function, and let
$a_{t,i}, r_{t,j} \in \{0,1\}$ denote pass/fail outcomes.

\begingroup
\small
\begin{equation}
\begin{aligned}
\mathrm{ACR} &=
\frac{\sum_{t=1}^{N}\sum_{i=1}^{A_t} a_{t,i}}
     {\sum_{t=1}^{N} A_t},
&
\mathrm{CCR} &=
\frac{\sum_{t=1}^{N}\sum_{j=1}^{R_t} r_{t,j}}
     {\sum_{t=1}^{N} R_t},\\
\mathrm{IPR} &=
\frac{\sum_{t=1}^{N}\left(\sum_{i=1}^{A_t} a_{t,i} + \sum_{j=1}^{R_t} r_{t,j}\right)}
     {\sum_{t=1}^{N}(A_t + R_t)},
&
\mathrm{FCCR} &=
\frac{1}{N}\sum_{t=1}^{N}
\mathbb{I}\!\left[
\left(\sum_{i=1}^{A_t} a_{t,i} = A_t\right)
\wedge
\left(\sum_{j=1}^{R_t} r_{t,j} = R_t\right)
\right].
\end{aligned}
\end{equation}
\endgroup

Here ACR and CCR measure atomic and relation-item satisfaction, IPR aggregates
all checklist items, and literal all-item FCCR requires every item in a task to pass. ACR
isolates compliance with individual requirements, while CCR captures the
harder ability to coordinate multiple requirements under \texttt{AND},
\texttt{THEN}, and \texttt{PRIORITY} relations; FCCR is the strictest
checklist-level summary. We report aggregate metrics, category-level ACR,
and relation-level CCR.

\paragraph{Pooled aggregation.}
Let $A=\sum_t A_t$ and $R=\sum_t R_t$. The pooled definitions imply
\begin{equation}
\mathrm{IPR}=\frac{A\,\mathrm{ACR}+R\,\mathrm{CCR}}{A+R}.
\label{eq:ipr_aggregation}
\end{equation}
For ANCHOR, $A=2{,}836$ and $R=600$. System-level IPR is computed according to Equation~\ref{eq:ipr_aggregation} over the common item set. All metrics are reported to one decimal place.

Category-level \emph{Macro ACR} instead averages the per-task category ACR
over task--configuration pairs containing at least one atomic item from
that category. It assigns equal weight to each eligible pair, whereas
\emph{Pooled ACR} weights each atomic item equally. We label these
aggregations explicitly in tables and figures.

\paragraph{Interpreting strict compliance.}
FCCR answers whether every scored checklist item passes; it is neither a
severity-weighted risk score nor the rate of jointly successful and compliant
task completions. A missed report item and an API-breaking change both make
FCCR zero for a task, although their engineering impact differs. ACR, CCR,
and IPR retain partial-compliance information, and task completion must be
reported alongside them. FCCR also depends on checklist length and the
dependence between items, so a lower value on longer checklists alone does
not establish a coordination deficit.

\paragraph{Priority and applicability.}
A \texttt{PRIORITY} relation can pass when the higher-priority requirement
is met and the lower-priority requirement is sacrificed. A failed
lower-priority atomic still makes FCCR zero. FCCR therefore measures literal
checklist completion, rather than satisfaction of obligations after resolving
precedence. A precedence-aware alternative would require explicit applicability
labels for overridden obligations and a separate scoring rule.

\paragraph{Weighting and category imbalance.}
The equations above pool checklist items: frequent categories and tasks with
more items receive more weight in ACR/IPR. This matters because C2, C3, and
C9 contain 1,560 of the 2,836 atomics (55.0\%), while C6--C8 contain only
168 (5.9\%). Their frequencies reflect this task collection and generation
pipeline, not an independently estimated distribution of engineering needs.
Possible complements are a macro-average of category-specific ACR values
and a weighted item-pass rate
$\sum_{t,k} w_{t,k}y_{t,k}/\sum_{t,k}w_{t,k}$, where $y_{t,k}$ is an
atomic or relation-item outcome and weights are fixed before comparing
systems. Such weights could reflect a documented deployment's priorities;
they would not turn strict all-or-nothing FCCR into a severity-sensitive
metric. These alternatives are not included in the reported scores.

\section{Experiments}
\label{sec:experiments}

After summarizing the evaluation setup, we report three layers of findings.
First, ANCHOR exposes a large and systematic gap between task completion and
engineering constraint compliance. Second, this gap is structured rather than
uniform: it concentrates in trajectory-grounded categories and in composite
constraints that require temporal or relational coordination. Third, trajectory-based error attribution most often labels failures as
missed or deprioritized constraints. These labels provide descriptive
hypotheses about behavior, rather than a causal separation of model
capability and scaffold effects.

\subsection{Experimental Setup}
\label{sec:experiments:setup}

ANCHOR contains 358 tasks from 23 repositories spanning eight language
groups (counting C and C++ together), with 2,836 atomic constraints and 600 relation items across ten
top-level categories and 78 leaves. We distinguish three evaluation sets.
The \emph{primary paired set} contains seven foundation models evaluated
under both OpenCode and Cline, yielding 14 configurations:
DeepSeek-V3.2\citep{liu2025deepseek}, GLM-5\citep{zeng2026glm},
GLM-5.1\citep{zai2026glm51}, GPT-5.3-Codex\citep{openai2026gpt53codex},
GPT-5.4\citep{openai2026gpt54}, Kimi-K2.5\citep{team2026kimi}, and
Qwen-3.5-Plus. The \emph{extended paired set} also includes GPT-5.4-Mini
and Qwen-3.5-Flash under both scaffolds, yielding 18 configurations.
The \emph{full evaluation set} includes these 18 configurations and Claude
Opus 4.8 under Cline, for 19 configurations spanning ten models.
All configurations use the same 358 tasks. Table~\ref{tab:main_results} and
the system-level scatter cover the full evaluation set; diagnostic figures
specify the corresponding paired set in their captions.

\begin{figure}[!htbp]
\centering
\safeincludegraphics[width=\linewidth]{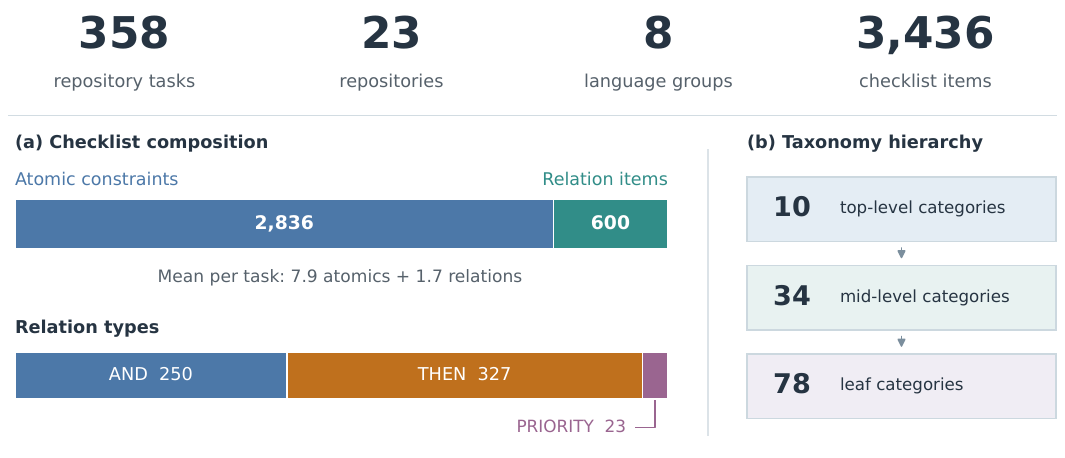}
\caption{ANCHOR dataset overview. Stacked-bar widths are proportional to item
counts; the relation-type bar expands the 600 relation items. Per-task counts
are means rounded to one decimal place. The taxonomy spans ten top-level
categories, 34 mid-level categories, and 78 leaves, all represented in the
benchmark. C and C++ are counted as one language group.}
\label{fig:dataset_overview}
\end{figure}

Each run starts from a clean repository checkout inside an isolated Docker
sandbox. Agents may inspect files, execute commands, and edit the repository
through their scaffold interfaces; network access is restricted to dependency
resolution required by the original environment. Task completion is evaluated
with the original repository test harness under a fixed 30-minute timeout, and
we record the full interaction trace for constraint evaluation. We report task
completion together with the four ANCHOR metrics from
Section~\ref{sec:evaluation_framework}: ACR, CCR, IPR, and FCCR.
Table~\ref{tab:main_results} reports point estimates; the category
plot reports task-bootstrap 95\% confidence intervals. These intervals
address variation across sampled tasks, not stochastic variation across
repeated agent runs. To separate constraint following from raw task
solvability, we additionally evaluate a \emph{task-only} setting in which the
same agent receives the repository task objective without the ANCHOR
engineering constraints. Benchmark statistics are summarized in
Figure~\ref{fig:dataset_overview}.

\FloatBarrier
\subsection{Main Results: The Constraint--Completion Gap}
\label{sec:experiments:main-results}

Table~\ref{tab:main_results} summarizes task completion and constraint
compliance for all evaluated agent configurations. The main result is a
consistent gap between producing a functionally correct patch and following the
engineering constraints that govern how the patch should be produced.

\begin{table}[!htbp]
\centering
\caption{
Main results on ANCHOR. Task-only completion is measured on the original task
objective without engineering constraints. Constrained completion is measured
with the full ANCHOR developer instruction. ACR, CCR, IPR, and FCCR are reported
on constrained runs. All numbers are percentages; higher is better.
ACR, CCR, and IPR pool checklist items; FCCR uses the literal all-item criterion.
\colorbox{goldbg}{\textbf{Gold}}, \colorbox{silverbg}{silver}, and
\colorbox{bronzebg}{bronze} cells mark the best, second-best, and third-best
values in each column.
}
\label{tab:main_results}
\renewcommand{\arraystretch}{1.15}
\resizebox{\textwidth}{!}{
\begin{tabular}{llcccccc}
\toprule
\rowcolor{headerbg}
\textbf{Scaffold} & \textbf{Model} & \multicolumn{2}{c}{\textbf{Task Completion}}
& \multicolumn{4}{c}{\textbf{Constraint Compliance}} \\
\cmidrule(lr){3-4}\cmidrule(lr){5-8}
\rowcolor{headerbg}
& & \textbf{Task-only} & \textbf{Constrained}
& \textbf{ACR} & \textbf{CCR} & \textbf{IPR} & \textbf{FCCR} \\
\midrule
\multirow{9}{*}{OpenCode}
& DeepSeek-V3.2 & 24.3 & 23.2 & 40.7 & 28.3 & 38.5 & 1.7 \\
& GLM-5 & 31.8 & 33.5 & 59.2 & 52.5 & 58.0 & 4.2 \\
& GLM-5.1 & \bronzec{36.0} & 35.5 & 60.6 & 57.3 & 60.0 & 6.1 \\
& GPT-5.3-Codex & 35.5 & \bronzec{37.4} & \silverc{68.9} & \silverc{63.4} & \silverc{67.9} & \silverc{12.3} \\
& GPT-5.4 & 34.9 & 36.3 & \bronzec{67.4} & \bronzec{62.6} & \bronzec{66.6} & 11.5 \\
& Kimi-K2.5 & 27.4 & 29.3 & 57.6 & 50.2 & 56.3 & 4.2 \\
& Qwen-3.5-Plus & 28.2 & 32.7 & 52.6 & 47.5 & 51.7 & 2.5 \\
& GPT-5.4-Mini & 31.2 & 30.6 & 60.1 & 51.1 & 58.5 & 7.3 \\
& Qwen-3.5-Flash & 21.5 & 18.7 & 35.7 & 27.4 & 34.3 & 1.0 \\
\midrule
\multirow{10}{*}{Cline}
& DeepSeek-V3.2 & 27.6 & 23.5 & 51.2 & 38.5 & 49.0 & 2.5 \\
& GLM-5 & 33.2 & 33.2 & 61.4 & 56.7 & 60.6 & 5.6 \\
& GLM-5.1 & 34.9 & 33.8 & 61.2 & 57.0 & 60.5 & 7.8 \\
& GPT-5.3-Codex & \silverc{36.9} & \silverc{37.7} & 64.1 & 57.5 & 62.9 & 9.2 \\
& GPT-5.4 & 33.5 & 36.9 & 64.3 & 60.9 & 63.7 & \bronzec{12.0} \\
& Kimi-K2.5 & 32.7 & 31.0 & 59.4 & 57.9 & 59.1 & 3.9 \\
& Qwen-3.5-Plus & 29.3 & 32.1 & 45.9 & 40.8 & 45.0 & 2.2 \\
& GPT-5.4-Mini & 23.1 & 22.3 & 32.4 & 23.5 & 30.8 & 4.1 \\
& Qwen-3.5-Flash & 24.9 & 22.7 & 50.4 & 39.2 & 48.4 & 2.8 \\
& Claude Opus 4.8 & \goldc{45.3} & \goldc{43.6} & \goldc{70.1} & \goldc{63.7} & \goldc{69.0} & \goldc{19.0} \\
\bottomrule
\end{tabular}
}
\end{table}

Claude Opus 4.8 under Cline achieves the highest reported compliance scores:
70.1\% ACR, 63.7\% CCR, 69.0\% IPR, and 19.0\% FCCR, with 43.6\%
constrained task completion. Its literal all-item FCCR is 7.0 percentage
points higher than GPT-5.4 under Cline (12.0\%). Nevertheless, 81.0\% of tasks still fail at least one checklist item.
Completion and FCCR are different marginal rates, not a measured joint
success rate; their 24.6-point difference lower-bounds the share of tasks
that are completed without passing the entire checklist. The strict
interpretation of FCCR, including overridden lower-priority atomics, is
specified in Section~\ref{sec:eval:metrics}.

The task-only comparison is a calibration check. Across the primary
paired set, the mean \emph{signed} completion change is $+0.7$
percentage points, while the mean absolute change is 1.9 points; individual
changes range from $-4.1$ to $+4.5$ points. For Claude Opus 4.8, completion
falls from 45.3\% to 43.6\%, a $-1.7$-point change. Across all 19 rows
with task-only values in Table~\ref{tab:main_results}, the mean signed
change is approximately $+0.1$ percentage points. These differences do not establish equivalence of task difficulty or joint satisfiability of constraints.

The compliance gap also persists under the same foundation model across
scaffolds. OpenCode+GPT-5.3-Codex and Cline+GPT-5.3-Codex differ by only
0.3 points in constrained completion (37.4\% versus 37.7\%), but by
5.0 points in IPR (67.9\% versus 62.9\%) and 3.1 points in FCCR
(12.3\% versus 9.2\%). Thus similar functional performance can coexist
with materially different compliance profiles.

The metric ordering is itself diagnostic. Agents satisfy isolated atomic items
substantially more often than relation items, with ACR exceeding CCR by 6.7
percentage points on average across the 19 configurations. FCCR is much lower than both because it is a strict
task-level metric: a single missed checklist item causes full compliance to
fail. The four metrics therefore distinguish average item satisfaction
from passing every scored item jointly, subject to the severity and priority
limitations discussed in Section~\ref{sec:eval:metrics}.

\subsection{Fine-Grained Results by Constraint Category}
\label{sec:experiments:taxonomy}
\label{sec:exp:category}

\begin{figure}[!htbp]
\centering
\safeincludegraphics[width=0.98\linewidth]{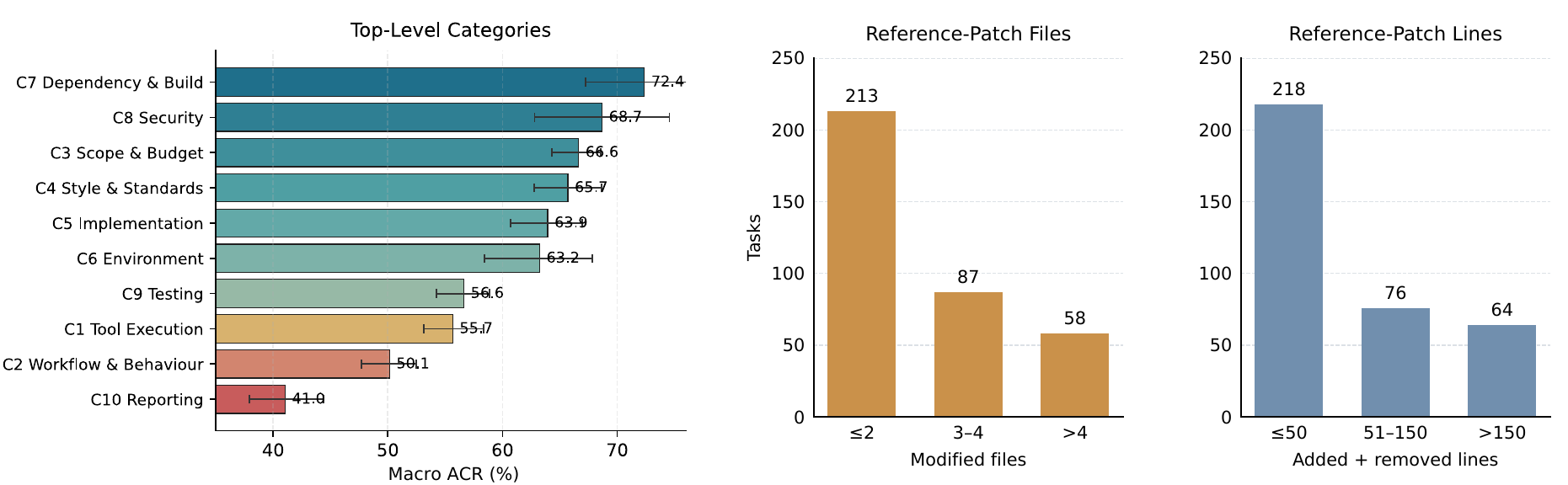}
\caption{Left: category-level Macro ACR for the primary paired set.
Rates are averaged over task--configuration pairs containing at least one
atomic item from the corresponding category. Error bars denote 95\%
bootstrap confidence intervals over tasks. Middle and right: reference-patch
scope distributions over all 358 tasks. File counts include every modified
file; changed lines count additions plus deletions within diff hunks.
Each scope panel contains one observation per task.}
\label{fig:category_complexity}
\end{figure}

\begin{figure}[!htbp]
\centering
\safeincludegraphics[width=0.98\textwidth]{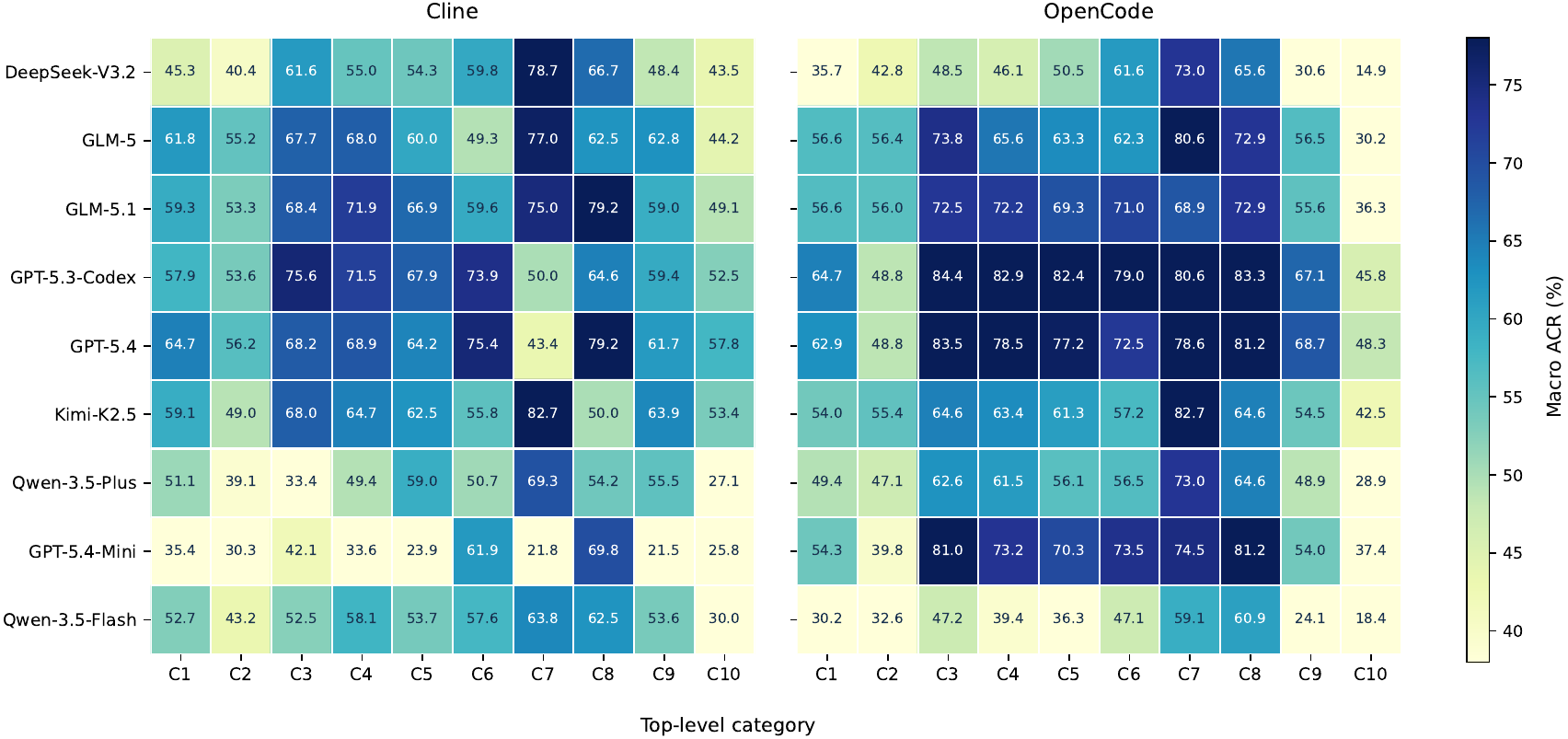}
\caption{
Per-configuration top-level category breakdown. Each cell reports Macro ACR
(\%) for one scaffold--model configuration and one top-level category,
macro-averaged over tasks containing at least one atomic constraint from that
category. Columns use the C1--C10 shorthand from
Table~\ref{tab:taxonomy-top}. The heatmap covers the extended paired
set (18 configurations). It makes it easier to identify
whether a category-level trend in Figure~\ref{fig:category_complexity} is
broad-based or driven by a subset of configurations.
}
\label{fig:category_model_heatmap}
\end{figure}

Figure~\ref{fig:category_complexity} shows that category difficulty is highly
non-uniform, and the pattern aligns with the kind of behavior each category
demands. The highest point estimates are for Dependency \& Build Strategy
(72.4\% Macro ACR), Security \& Dangerous Operations (68.7\%), and Modification
Scope \& Change Budget (66.6\%). These constraints are often phrased as local
prohibitions or boundary conditions, so conservative agent behavior is
frequently sufficient to satisfy them. In contrast, Output Format \& Reporting
is by far the hardest category (41.0\% Macro ACR), followed by Workflow \&
Behavior (50.1\%), Tool Execution Protocol (55.7\%), and Testing (56.6\%).
These categories are more trajectory-grounded: they require agents to maintain
process discipline over time, choose the right tools, act in the correct order,
validate the result, and faithfully report what they actually did.
Appendix~\ref{app:taxonomy_leaf} documents the taxonomy at mid-level
and leaf-level granularity. Figure~\ref{fig:category_model_heatmap} shows
per-configuration Macro ACR for the extended paired set.

The scope distributions describe the reference patches associated with
the tasks: 213 modify at most two files, 87 modify three or four, and 58
modify more than four. By changed lines, 218 contain at most 50 additions
plus deletions, 76 contain 51--150, and 64 contain more than 150.

The rare categories need particular caution: C6 has 70 atomic
constraints, and C7 and C8 have only 49 each, compared with 512--530 in
C2, C3, and C9. Re-evaluating the same items across models does not create
new independent tasks. The displayed point estimates therefore do not
establish reliable superiority on security or dependency handling; category
coverage and judging methods are detailed in
Table~\ref{tab:category_evidence}. Category-specific IPR/FCCR are
not reported because relation items may span multiple atomic categories.

\paragraph{Category coverage and judging reliability.}
\label{app:category_reliability}
Table~\ref{tab:category_evidence} combines the category Macro ACR estimates
from Figure~\ref{fig:category_complexity}, atomic and distinct-task counts, and applicable
evaluation methods. These estimates use the figure's task--configuration
macro-average, distinct from pooling all atomic items. Methods are selected
for individual pass conditions: no category is exclusively rule-judged or
LLM-judged. In particular, \texttt{import\_check} tests named imports, not
arbitrary dependency changes; \texttt{action\_sequence} tests observable
action sequences, not whether an agent semantically understood a prerequisite.

The judge study reports $\kappa=0.86$ for rule-verifiable items and
$\kappa=0.69$ for difficult \texttt{THEN}/\texttt{PRIORITY} items.
These pooled coefficients cannot be assigned to every category in the
corresponding group. The 50-instance audit likewise does not provide
per-category validity estimates. Category-specific IPR/FCCR would require
an explicit attribution rule for cross-category relation items, whose
category fields are null in the checklist schema; they are not reported
here. Future uncertainty estimates for these breakdowns should resample
whole tasks and retain within-task dependence. The 49-item C7/C8 and 70-item C6 slices warrant cautious,
directional interpretation.

\begin{table}[!htbp]
\centering
\caption{Category counts, Macro ACR (\%), and typical evidence and
checking methods. Macro ACR averages over task--configuration pairs
containing the corresponding category. The Tasks column counts distinct instances
containing at least one such atomic item. Category names appear in
Table~\ref{tab:taxonomy-top}.}
\label{tab:category_evidence}
\small
\begin{tabular}{lrrrL{7.7cm}}
\toprule
Category & Atomics & Tasks & \shortstack{Macro\\ACR} & Typical evidence and checking methods \\
\midrule
C1 & 371 & 234 & 55.7 & Command presence/argument rules; semantic result verification may need an LLM. \\
C2 & 512 & 284 & 50.1 & Observable action sequences; semantic ordering, explanations, and precedence need an LLM. \\
C3 & 518 & 274 & 66.6 & Path/file/line checks; semantic scope and API compatibility need deeper inspection. \\
C4 & 223 & 167 & 65.7 & Diff patterns and lint evidence; naming, documentation quality, and design need semantic judgment. \\
C5 & 283 & 174 & 63.9 & Named imports and code patterns; algorithms, error handling, and side effects need semantic judgment. \\
C6 & 70 & 66 & 63.2 & Paths, commands, and configuration; runtime compatibility may require execution or semantic evidence. \\
C7 & 49 & 47 & 72.4 & Manifest paths and named imports; general dependency/build policy may need an LLM. \\
C8 & 49 & 48 & 68.7 & Prohibited commands and credential patterns; secure implementation needs semantic checks. \\
C9 & 530 & 297 & 56.6 & Test commands, flags, and ordering; test adequacy and result interpretation need semantic judgment. \\
C10 & 231 & 162 & 41.0 & Report text/patterns; completeness and fidelity need comparison with trajectory evidence. \\
\bottomrule
\end{tabular}
\end{table}

\subsection{Scaffold--Model Interactions}
\label{sec:experiments:scaffolds}
\label{app:scaffold_comparison}
For the same foundation model, IPR can differ substantially between
OpenCode and Cline. GPT-5.4-Mini scores 58.5\% under OpenCode and 30.8\%
under Cline, a decrease of 27.7 points; DeepSeek-V3.2 and Qwen-3.5-Flash
score 10.5 and 14.1 points higher under Cline, respectively. The strongest
paired configuration also depends on the scaffold: GPT-5.3-Codex has the
highest IPR under OpenCode, whereas GPT-5.4 leads the paired Cline
configurations. Table~\ref{tab:scaffold_differences} reports per-model
point differences for the extended paired set. These descriptive
comparisons cover complete scaffold configurations, including prompts,
tools, and workflow; they do not isolate a causal component or establish
statistical significance. We use two open-source scaffolds to make their
interfaces inspectable. Performance under Claude Code, Codex, or other
scaffolds is outside the evaluated scope; Claude Opus 4.8 is evaluated
under Cline only.

\begin{table}[!htbp]
\centering
\caption{Scaffold differences (Cline minus OpenCode) in percentage points.}
\label{tab:scaffold_differences}
\small
\begin{tabular}{lrrrr}
\toprule
Model & $\Delta$ACR & $\Delta$CCR & $\Delta\mathrm{IPR}$ & $\Delta$FCCR \\
\midrule
DeepSeek-V3.2 & $+10.5$ & $+10.2$ & $+10.5$ & $+0.8$ \\
GLM-5 & $+2.2$ & $+4.2$ & $+2.6$ & $+1.4$ \\
GLM-5.1 & $+0.6$ & $-0.3$ & $+0.5$ & $+1.7$ \\
GPT-5.3-Codex & $-4.8$ & $-5.9$ & $-5.0$ & $-3.1$ \\
GPT-5.4 & $-3.1$ & $-1.7$ & $-2.9$ & $+0.5$ \\
Kimi-K2.5 & $+1.8$ & $+7.7$ & $+2.8$ & $-0.3$ \\
Qwen-3.5-Plus & $-6.7$ & $-6.7$ & $-6.7$ & $-0.3$ \\
GPT-5.4-Mini & $-27.7$ & $-27.6$ & $-27.7$ & $-3.2$ \\
Qwen-3.5-Flash & $+14.7$ & $+11.8$ & $+14.1$ & $+1.8$ \\
\bottomrule
\end{tabular}
\end{table}

\subsection{Composite Constraints: Semantics and Structural Difficulty}
\label{sec:experiments:composite}

We next isolate the structured part of engineering constraint following:
whether agents preserve the relations linking multiple constraints, not just the
atomic requirements themselves.

\begin{figure}[!htbp]
\centering
\safeincludegraphics[width=0.93\textwidth]{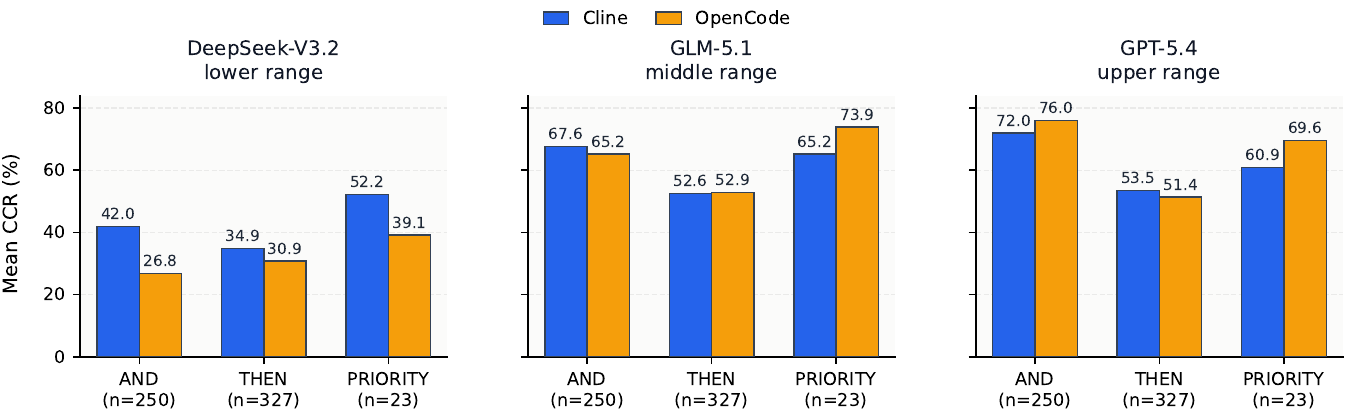}
\caption{Semantic breakdown of composite relation types for three models under both scaffolds.}
\label{fig:relation_type_breakdown}
\end{figure}

\begin{figure}[!htbp]
\centering
\safeincludegraphics[width=0.98\textwidth]{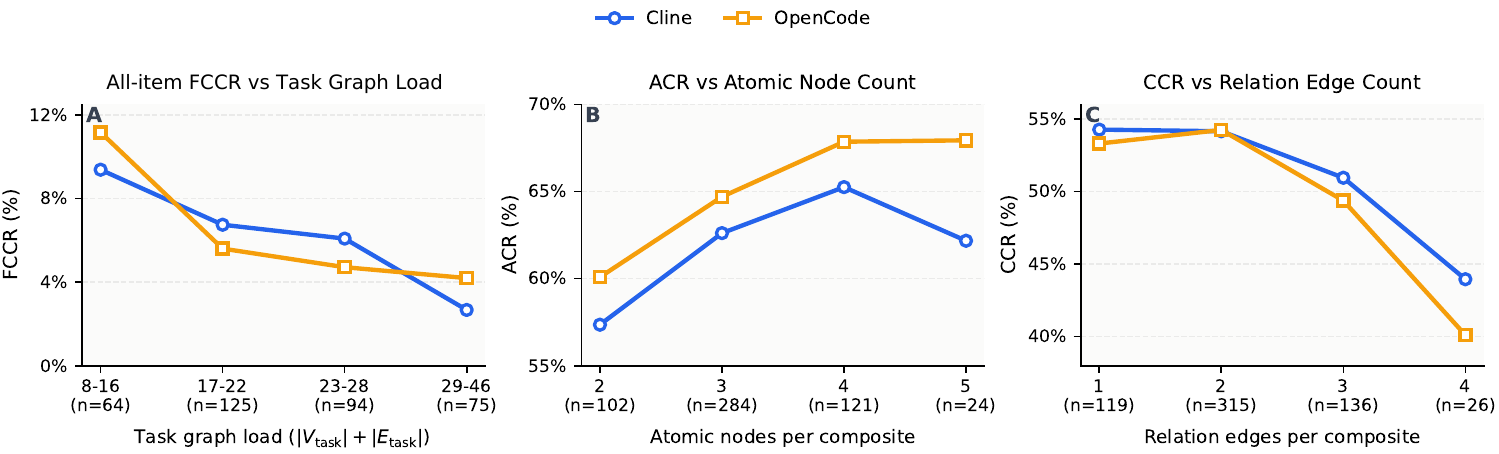}
\caption{
Structural diagnosis of composite constraints. From left to right: FCCR as a
function of task-level graph load, computed on the union graph induced by all
composite constraints in a task; ACR as a function of the number of atomic
nodes in a composite; and CCR as a function of the number of relation edges in
a composite. Each scaffold curve averages the seven models in the primary paired set
under that scaffold. Sample counts are shown in the x-axis tick
labels. The displayed counts sum to 358 tasks in the graph-load
panel, 531 observations grouped by composite node count, and 596 grouped
by composite edge count. The latter two are panel-specific observation
counts, distinct from the benchmark-wide atomic and relation-item totals. The ACR and CCR panels use
truncated y-axes for readability.
}
\label{fig:composite_structure}
\end{figure}

\begin{figure}[!htbp]
\centering
\safeincludegraphics[width=0.90\textwidth]{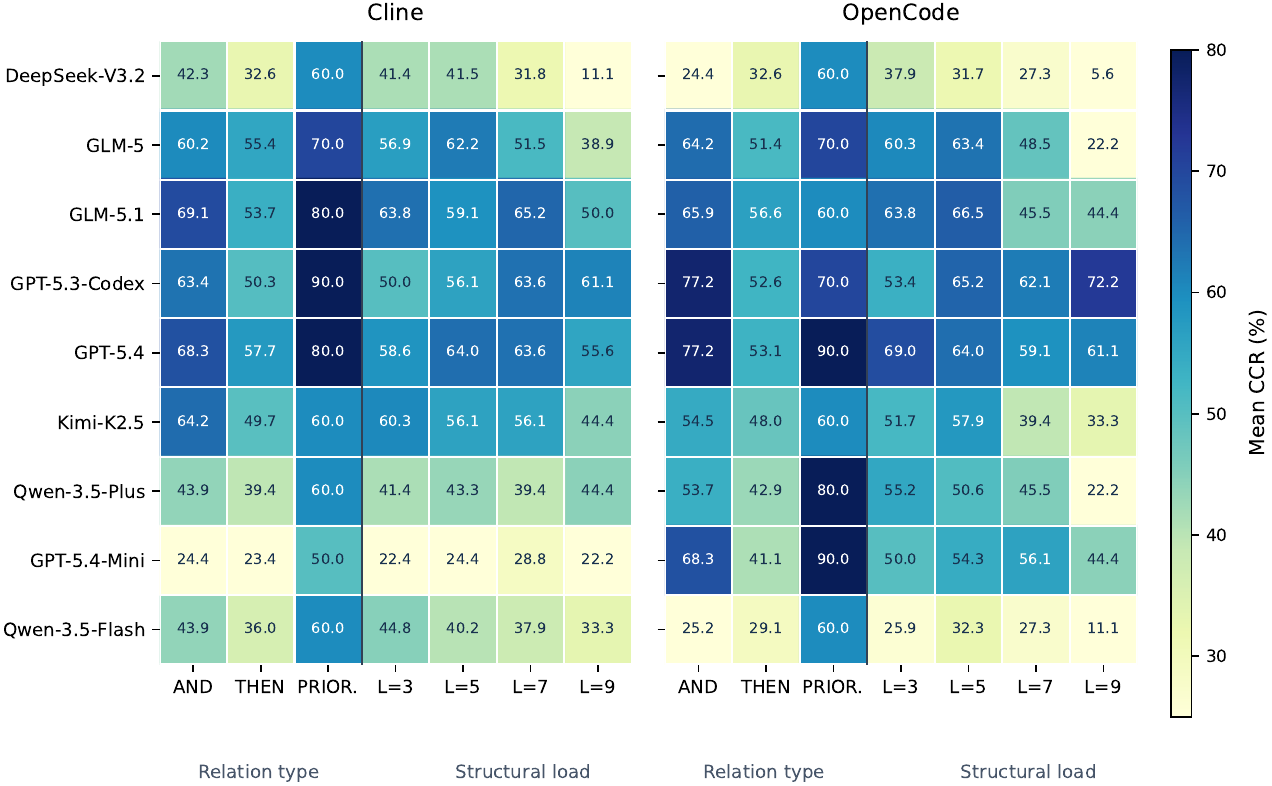}
\caption{
Per-configuration composite-constraint breakdown. Each cell reports mean CCR
(\%) for one scaffold--model configuration and one composite bucket. The first
three columns correspond to relation types (\texttt{AND}, \texttt{THEN},
\texttt{PRIORITY}); the remaining columns correspond to retained structural-load
bins, where load is defined as $|V|+|E|$. The heatmap shows that the
\texttt{THEN} penalty is broad-based across the extended paired set,
while performance on higher-load
composites is much more model-sensitive.
}
\label{fig:composite_model_heatmap}
\end{figure}

Figure~\ref{fig:relation_type_breakdown} isolates relation semantics.
\texttt{THEN} is clearly the hardest relation type, with mean CCR only 46.7\%,
compared with 58.1\% for \texttt{AND} and 64.0\% for \texttt{PRIORITY}. This
\texttt{THEN} penalty is broad-based across representative low-, mid-, and
high-performing model families, indicating that agents can often satisfy the
component actions yet still fail to preserve their required order. Failed
priority trajectories show a related pattern: agents optimize for the most
direct functional fix while neglecting the higher-priority engineering
requirement. The extended paired set is shown in
Figure~\ref{fig:composite_model_heatmap}.

Figure~\ref{fig:composite_structure} then turns to graph structure. The left
panel shows that the literal all-item pass rate decreases as the
overall task graph grows: FCCR drops from 9.4\% to 2.7\% on Cline and from
11.2\% to 4.2\% on OpenCode as task graph load increases from 8--16 to 29--46.
The middle panel asks whether this degradation is already visible at the level
of the local atomic requirements inside each composite. It is not. ACR stays
comparatively stable across 2--5-node composites, ranging from 57.4\% to 65.3\%
on Cline and from 60.1\% to 67.9\% on OpenCode. Larger
composites often bundle individually explicit local requirements, which helps
explain why ACR does not decline monotonically with node count. The right panel
shows where the main structural penalty enters: relation-level compliance drops
as edge count grows, from 54.3\% to 44.0\% on Cline and from 53.3\% to 40.1\%
on OpenCode across 1--4 edges. These associations are consistent with difficulty in coordinating
requirements, but do not isolate a causal effect of graph structure. The
strict FCCR criterion becomes harder as more items are included, and graph
size, relation type, and task difficulty are not independently randomized.
Count-controlled comparisons would be needed to separate these effects.

\subsection{Compliance and Task Completion Are Distinct}
\label{sec:experiments:orthogonality}
\label{app:conditional_compliance}

\begin{figure}[!htbp]
\centering
\safeincludegraphics[width=0.96\linewidth]{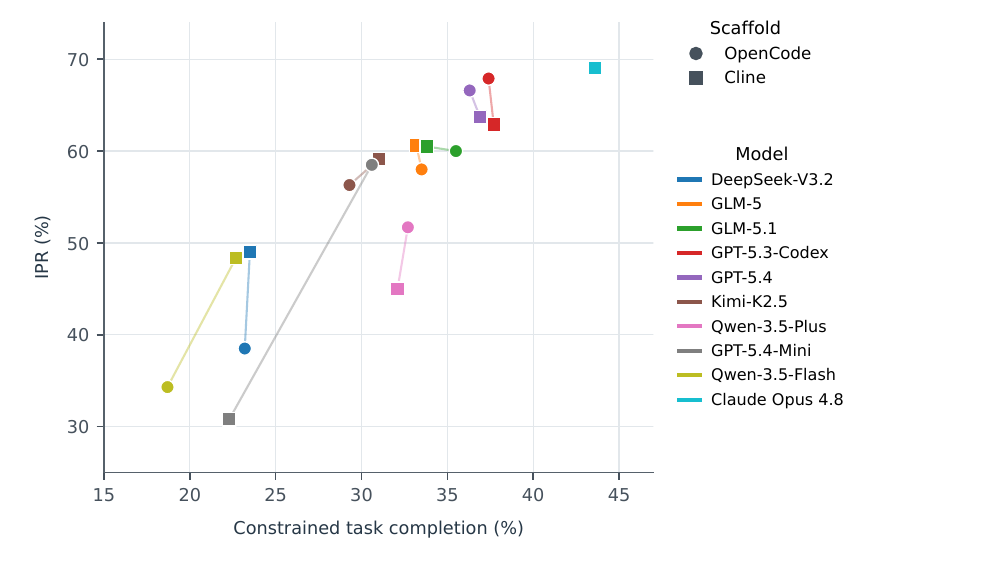}
\caption{Constrained task completion versus IPR for the full
evaluation set in Table~\ref{tab:main_results}. Colors identify models,
shapes identify scaffolds, and lines connect configurations using the same
foundation model across scaffolds. Each point represents a complete agent
configuration.}
\label{fig:compliance_vs_completion}
\end{figure}

Figure~\ref{fig:compliance_vs_completion} shows a positive association
between task completion and IPR across configurations. The Spearman
correlation is $0.87$ on the primary paired set, $0.90$ on the extended
paired set, and $0.91$ on the full evaluation set. Stronger configurations
tend to improve both metrics, although this system-level association does
not imply that the same tasks are easy under both criteria.

The gap between completion and literal all-item FCCR provides a complementary
comparison. For Claude Opus 4.8 under Cline, the marginal rates of 43.6\%
completion and 19.0\% FCCR imply that at least 24.6\% of tasks are completed
without passing every checklist item. This is a lower bound from marginal
rates, not a measured joint rate. As illustrated in Figure~\ref{fig:intro},
functional completion alone does not establish literal checklist completion.

Table~\ref{tab:success_failure_compliance} conditions FCCR on task success
for the extended paired set. FCCR is higher on successful trajectories in
13 of 18 configurations, with a mean signed difference of 1.74 percentage
points. These descriptive gaps do not establish statistical independence.
On failed tasks, a checklist pass does not imply a usable repair and may
include satisfied prohibitions or vacuously true conditional checks.

\begin{table}[!htbp]
\centering
\caption{Literal all-item FCCR on task-successful (S) and task-failed
(F) trajectories for the extended paired set (18 configurations).
All values are percentages.}
\label{tab:success_failure_compliance}
\small
\setlength{\tabcolsep}{6pt}
\renewcommand{\arraystretch}{1.12}
\begin{tabular}{lrrrr}
\toprule
\textbf{Model} & \multicolumn{2}{c}{\textbf{OpenCode}} & \multicolumn{2}{c}{\textbf{Cline}} \\
\cmidrule(lr){2-3}\cmidrule(lr){4-5}
& \textbf{FCCR (S)} & \textbf{FCCR (F)} & \textbf{FCCR (S)} & \textbf{FCCR (F)} \\
\midrule
DeepSeek-V3.2 & 4.8 & 0.7 & 6.0 & 1.5 \\
GLM-5 & 6.7 & 2.9 & 4.2 & 6.3 \\
GLM-5.1 & 3.9 & 7.4 & 7.4 & 8.0 \\
GPT-5.3-Codex & 11.9 & 12.5 & 11.9 & 7.6 \\
GPT-5.4 & 13.8 & 10.1 & 12.9 & 11.5 \\
Kimi-K2.5 & 5.7 & 3.6 & 5.4 & 3.2 \\
Qwen-3.5-Plus & 5.1 & 1.2 & 5.2 & 0.8 \\
GPT-5.4-Mini & 9.1 & 6.5 & 6.2 & 3.6 \\
Qwen-3.5-Flash & 1.5 & 1.0 & 1.2 & 3.2 \\
\bottomrule
\end{tabular}
\end{table}

\FloatBarrier
\subsection{Error Analysis}
\label{sec:experiments:error_analysis}

To understand where constraint failures come from, we analyze the attributed
error labels produced by our failure-analysis pipeline over the primary paired set. This yields 20,726 attributed failures in
total: 16,700 from atomic checklist items and 4,026 from composite relation
items. We group them into six attributed behavioral patterns: \textbf{E1} Not Perceived,
\textbf{E2} Misunderstood, \textbf{E3} Deprioritized, \textbf{E4} Capability
Limitation, \textbf{E5} Scaffold Interference, and \textbf{E6} Partial
Compliance. The attribution prompt used in this step is provided in
Appendix~\ref{app:prompts:error}.

\begin{figure}[!htbp]
\centering
\safeincludegraphics[width=0.96\textwidth]{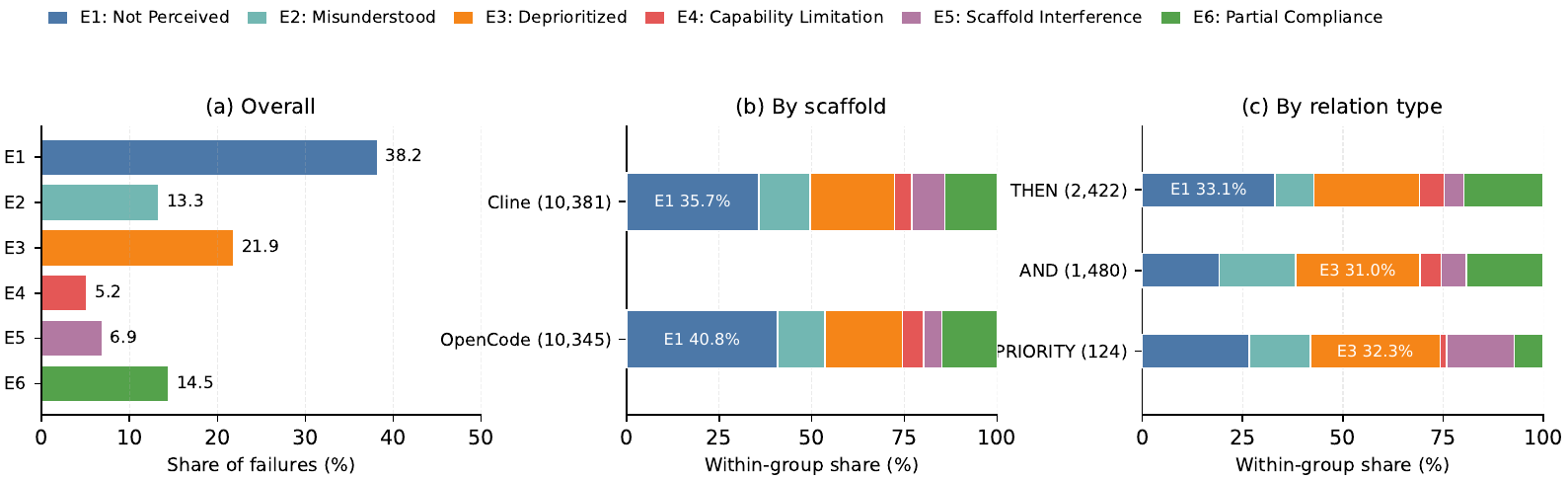}
\caption{
Attributed behavioral patterns across failed ANCHOR checklist items in the
primary paired set. Left: overall distribution over 20,726
failures. Middle: row-normalized composition by scaffold. Right:
row-normalized composition by relation type for 4,026 failed composite items.
E1 = not perceived; E2 = misunderstood; E3 = deprioritized; E4 = capability
limitation; E5 = scaffold interference; E6 = partial compliance.
}
\label{fig:error_attribution_summary}
\end{figure}

Figure~\ref{fig:error_attribution_summary} summarizes the attributed behavioral patterns.
E1 \emph{Not Perceived} accounts for 38.2\% of failures, E3
\emph{Deprioritized} for 21.9\%, and E6 \emph{Partial Compliance} for
14.5\%; together they account for 74.6\%, while E4 \emph{Capability
Limitation} accounts for 5.2\%. These are post-hoc labels inferred from
constraint text and judge reasoning, not direct observations of the model's
internal state or experimentally identified causes.

E5 \emph{Scaffold Interference} accounts for 6.9\% of attributed failures
and covers recognizable interference such as context truncation, conflicting
system instructions, and tool/API limitations. This share does not bound
the scaffold's total effect: different prompts, tools, and workflows can
change whether a model notices or prioritizes a constraint, yielding behavior
labelled E1 or E3. Thus a small E5 share is compatible with the large
scaffold--model interactions in Section~\ref{sec:experiments:scaffolds}.
The attribution data alone do not quantify those indirect pathways.

\texttt{THEN} failures skew toward E1/E6, whereas \texttt{AND} and
\texttt{PRIORITY} failures more often receive E3. These patterns motivate
better workflow tracking, persistent constraint-state maintenance, and
precedence reasoning as directions to test, alongside improvements in
coding capability and scaffold design.

\begin{figure}[!htbp]
\centering
\safeincludegraphics[width=0.72\textwidth]{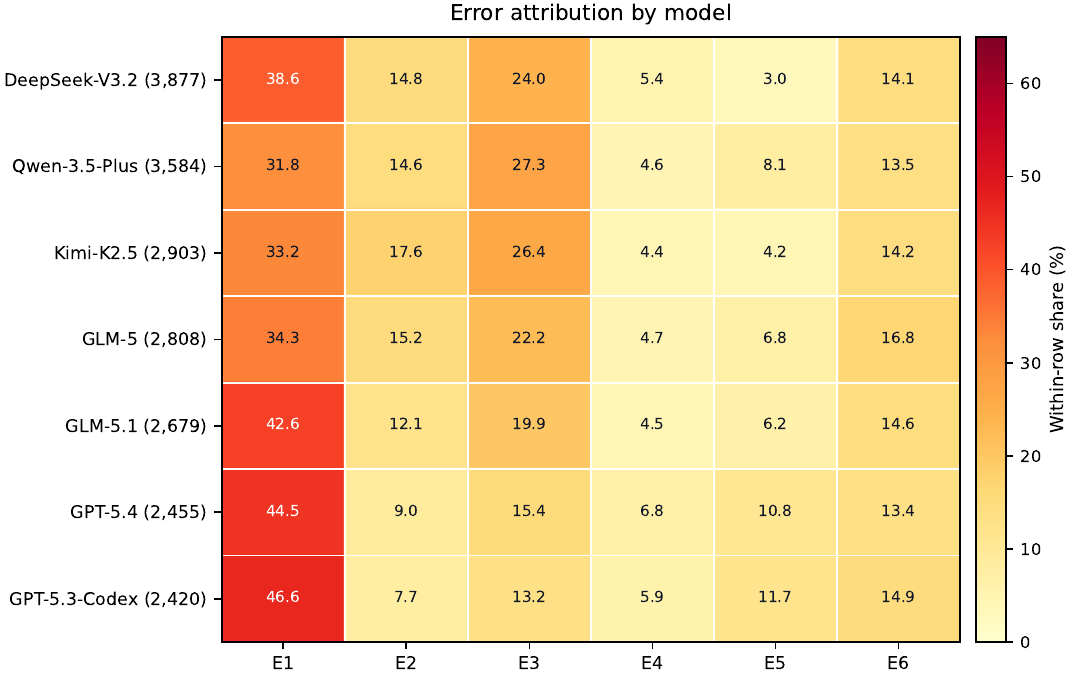}
\caption{
Row-normalized error attribution by model for the primary paired set. Rows
aggregate both scaffolds; cells show the percentage of attributed failures
assigned to E1--E6, and counts in parentheses show row totals. The labels
describe behavioral patterns, not causal effects of model capability.
}
\label{fig:error_attribution_model_heatmap}
\end{figure}

\begin{figure}[!htbp]
\centering
\safeincludegraphics[width=0.72\textwidth]{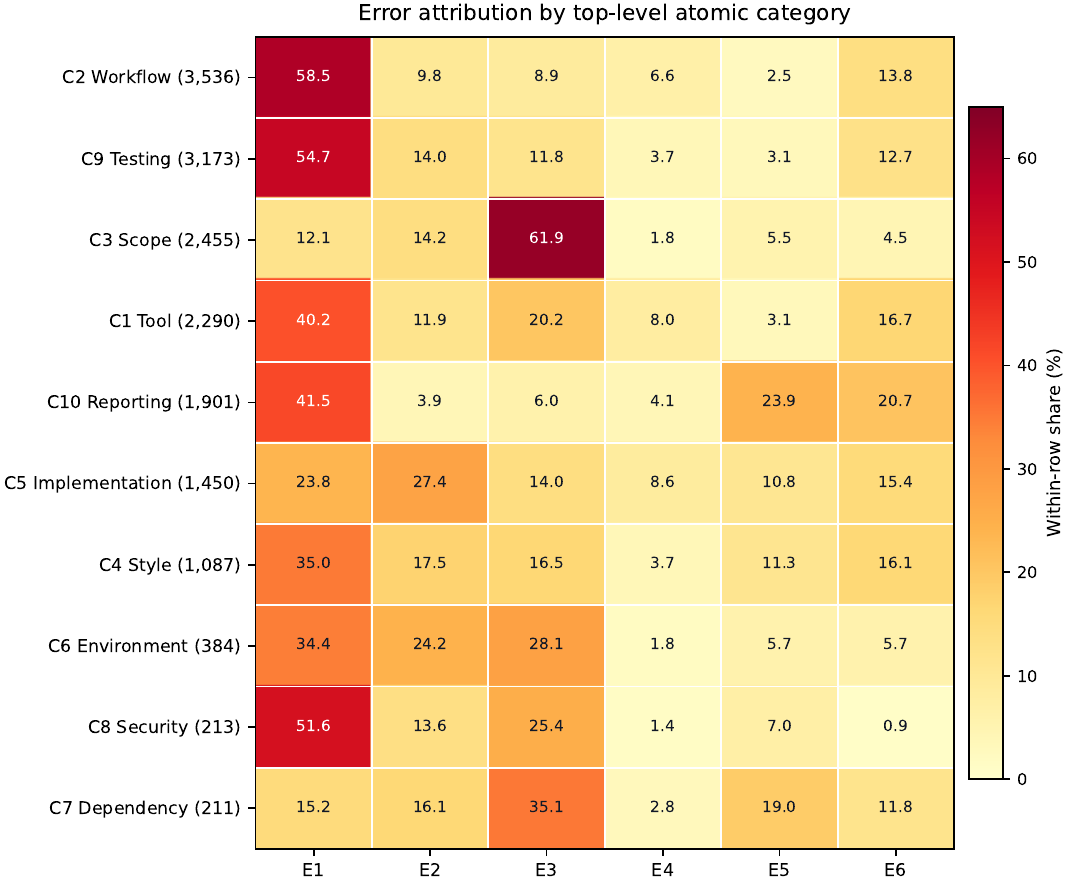}
\caption{
Row-normalized error attribution by atomic category. Cells show the percentage
of attributed failures assigned to E1--E6; counts in parentheses show row
totals. Workflow and Testing are dominated by E1, Modification Scope by E3,
and Output Format \& Reporting has a heavier mix of E5 and E6.
}
\label{fig:error_attribution_category_heatmap}
\end{figure}

Figures~\ref{fig:error_attribution_model_heatmap} and
\ref{fig:error_attribution_category_heatmap} break down the attributed
patterns by foundation model and constraint category.

\section{Discussion and Limitations}
\label{sec:discussion}

\subsection{Why Task Completion Is Not Deployment Readiness}
\label{sec:discussion:readiness}

Task completion is a necessary signal of coding ability, but it is not yet a
sufficient signal of deployment readiness. In real software engineering, the
acceptability of a patch depends not only on whether it resolves the underlying
issue, but also on whether it respects the engineering constraints under which
the task was assigned. Consider modification scope: an agent may fix a bug in
\texttt{parser.py}, yet also refactor adjacent modules despite an instruction to
keep changes local. The issue is resolved, but the patch becomes harder to
review, riskier to backport, and less likely to be merged. Likewise, an agent
may produce a functionally correct patch while violating interface constraints,
for example by changing a public API when explicitly asked to preserve backward
compatibility. In both cases, the artifact may satisfy a functional oracle while
failing the practical acceptance criteria imposed by the developer.

Process constraints create a second class of non-deployment-ready success. An
agent may arrive at a correct fix without first reproducing the failure, without
adding a regression test, or without reporting the commands it ran. Such a patch
is harder to audit, harder to maintain, and harder to trust in future incidents,
because reviewers lack evidence about how the failure was diagnosed and how the
fix was validated. Similarly, a task may be solved through prohibited means,
such as introducing an unapproved dependency, using a restricted tool, or
ignoring an explicit ordering requirement such as \emph{reproduce before fix} or
\emph{test before commit}. These examples illustrate why \emph{solve task} and
\emph{deployment-ready behaviour} are not equivalent in agentic coding. ANCHOR
is designed to make these otherwise hidden engineering failure modes measurable.
Section~\ref{sec:experiments:orthogonality} shows that system-level
completion and IPR are positively associated, while task completion and
literal all-item FCCR remain different outcomes. For the strongest
configuration, their marginal rates imply that approximately 24.6\% or
more of tasks are completed without passing every scored checklist item.
Table~\ref{tab:success_failure_compliance} also shows nonzero all-item
pass rates on task-failed trajectories.
Repository-level success and engineeringly acceptable behaviour therefore come
apart in both directions: agents can fix the issue while behaving like weak
collaborators, and they can behave in a largely disciplined way without
actually completing the repair.

\subsection{Limitations}
\label{sec:discussion:limitations}

\paragraph{Evaluation cost.}
Running each agent to completion on each ANCHOR task, and evaluating
its trajectory against the checklist, incurs
significant compute cost. This makes large-scale model sweeps, repeated judge
calibration, and continual benchmark refreshes substantially more expensive than
final-patch-only evaluation.

\paragraph{LLM judge reliability.}
Although hybrid judging broadens coverage, LLM-based evaluation may still share
systematic biases with the evaluated models, particularly for models from the
same vendor family as the judge. Future work should investigate cross-vendor
judge configurations and stronger human-audited judge calibration protocols.
The human alignment and subset-consistency checks do not substitute for
controlled judge-provider or prompt swaps. Error-attribution labels also rely
on LLM interpretation and should not be read as causal measurements.

\paragraph{Generated-constraint validity and reference-patch exposure.}
The five-engineer audit covers 50 instances rather than the full benchmark,
and evaluates atomics rather than complete interacting constraint sets.
The audit reports high raw agreement and Gwet's AC1. Prompt instructions against copying the reference
solution do not exclude indirect leakage, since both selection and
instantiation/composition receive the gold patch. A patch-blind construction
ablation and a full-set consistency audit remain open validation steps.

\paragraph{Strict metrics, coverage, and uncertainty.}
FCCR is sensitive to any failed item, irrespective of severity, and to
checklist size. Under the literal definition, it also penalizes a failed
lower-priority atomic even when the priority relation passes. An
applicability-aware variant needs explicit handling of overridden obligations
and new scoring. Category imbalance gives common constraints more weight in
pooled metrics; the small environment, dependency, and security slices do
not support definitive category rankings. Category-specific IPR/FCCR would
also require an attribution rule for relations spanning multiple categories.
Paired scaffold differences are descriptive comparisons without significance
tests or repeated-run uncertainty estimates.

\paragraph{Metric aggregation.}
ACR, CCR, and IPR pool scored checklist items, so tasks and categories with
more items receive proportionally greater weight. FCCR and category-level
Macro ACR expose complementary task-level and category-level behavior.
Applications may prefer other weighting schemes when particular engineering
requirements carry greater practical importance.

\paragraph{Scaffold scope.}
We evaluate two open-source scaffolds. Their prompts, tools, and workflows
can interact strongly with a given model, as the paired comparisons show.
Claude Opus 4.8 is evaluated under Cline only, and
the results cannot establish performance under untested scaffolds. The 6.9\%
E5 attribution share measures labelled interference, not the scaffold's total
causal contribution.

\paragraph{Taxonomy coverage.}
Our taxonomy was derived from grounded analysis of real developer interactions
with coding agents and the engineering requirements expressed in those tasks.
Despite our best efforts at coverage, niche engineering domains (e.g., embedded
systems or hardware description languages) may still contain constraint types
not captured in the current 78 leaf categories.

\paragraph{Constraint interaction effects.}
In ANCHOR, each composite constraint is treated as an independent subgraph. In
principle, constraints from different composites could interact---a PRIORITY
relationship between atomics in different composites is not currently modelled.
Disjoint graph membership does not imply semantic independence, and the
current prompt-based non-blocking checks do not prove joint satisfiability.
This is a known simplification.

\subsection{Benchmark Maintenance and Community Plan}
\label{sec:discussion:maintenance}

To ensure ANCHOR remains useful as models improve, we plan to maintain versioned
benchmark releases, contamination monitoring, and a documented expansion
protocol for adding new task instances under explicit quality-control
requirements. Published results should always cite the version used so that
future comparisons remain interpretable as the benchmark evolves.

\section{Conclusion}
\label{sec:conclusion}

We presented \textbf{ANCHOR}, a benchmark that reframes evaluation in agentic
coding from task completion alone to task completion under explicit engineering
constraints. By modelling task-level requirements as structured, often
trajectory-grounded constraints and evaluating them at multiple granularities,
ANCHOR makes \emph{engineering constraint following} measurable in
repository-grounded coding tasks. Across 19 configurations spanning two
scaffolds and ten models, the strongest measured system passes every scored
checklist item on 19.0\% of tasks, with 43.6\% constrained task completion.
Human audits provide evidence about sampled constraint validity and judge
agreement, while leaving uncertainty about full-set consistency and difficult
relational judgments. Compliance also depends substantially on scaffold--model
interactions. ANCHOR therefore supports evaluating task completion alongside
atomic, relational, and strict checklist compliance, with explicit attention
to priority semantics, category coverage, and evaluation uncertainty.

\FloatBarrier

\bibliographystyle{acl_natbib}
\bibliography{craftbench}

\appendix

\section{Full Constraint Taxonomy}
\label{app:taxonomy}
\label{app:taxonomy_leaf}

This appendix documents all 78 leaf categories in the current taxonomy. Across
the current release, these leaves account for 2,836 instantiated atomic
constraints. For each leaf, we provide:
(i) a one-sentence definition of what the constraint governs, (ii) 2--4
representative instantiations drawn from real developer instructions, and
(iii) the default evaluation mode (rule-based vs.\ LLM-as-judge).
Table~\ref{tab:taxonomy-top} summarizes the top-level taxonomy, while
Table~\ref{tab:taxonomy-leaf} provides the full mid- and leaf-level breakdown
used in the current release.

{\small\setlength{\tabcolsep}{4pt}
\begin{longtable}{L{2.45cm}L{3.15cm}cL{6.75cm}}
\caption{Mid- and leaf-level categories in the ANCHOR taxonomy, with instantiated counts and representative example constraints.}
\label{tab:taxonomy-leaf} \\
\toprule
\rowcolor{headerbg}
\textbf{Subcategory} & \textbf{Leaf category} & \textbf{Count} & \textbf{Example constraint} \\
\midrule
\endfirsthead

\toprule
\rowcolor{headerbg}
\textbf{Subcategory} & \textbf{Leaf category} & \textbf{Count} & \textbf{Example constraint} \\
\midrule
\endhead

\bottomrule
\endfoot

\rowcolor{headerbg}\multicolumn{4}{c}{\textbf{C1 Tool Execution Protocol}} \\
\midrule
1.1 Tool selection
& 1.1.1 Tool Preference \& Substitution & 44
& Use \texttt{rg} rather than \texttt{grep} for search; use \texttt{fd} rather than \texttt{find}. \\*
& 1.1.2 Tool prohibition & 1
& Do not use \texttt{sed -i} for bulk replacement; do not use \texttt{curl}. \\*
& 1.1.3 Tool Composition \& Pipeline & 1
& First locate the target with \texttt{rg}, then edit the file. \\
\midrule
1.2 Invocation parameters
& 1.2.1 Required parameters & 74
& Run \texttt{pytest} with \texttt{--timeout=30}. \\*
& 1.2.2 Forbidden parameters & 36
& Do not run \texttt{git commit} with \texttt{--no-verify}. \\
\midrule
1.3 Execution results
& 1.3.1 Truthfulness \& Completeness & 24
& Do not claim a command was executed unless it was actually run. \\*
& 1.3.2 Failure Handling \& Retry & 32
& Analyse the error before retrying a failed command. \\*
& 1.3.3 Result verification & 159
& Re-run the relevant command after editing to verify the effect. \\
\midrule

\rowcolor{headerbg}\multicolumn{4}{c}{\textbf{C2 Workflow \& Behaviour}} \\
\midrule
2.1 Execution order
& 2.1.1 Task-level order & 200
& Reproduce the bug before attempting a fix. \\*
& 2.1.2 Operation-level order & 9
& Read the file before modifying it. \\
\midrule
2.2 User confirmation
& 2.2.1 Pre-action confirmation & 1
& Present the fix plan and wait for approval before proceeding. \\*
& 2.2.2 Escalation triggers & 4
& Ask for confirmation before changing a public API signature. \\
\midrule
2.3 Information disclosure
& 2.3.2 Decision explanation & 87
& Explain why this fix is preferred over the alternatives. \\*
& 2.3.3 Capability Boundary Declaration & 9
& Mark uncertain conclusions as hypotheses rather than facts. \\
\midrule
2.4 Commit conventions
& 2.4.1 Commit message format & 21
& Use Conventional Commits format for commit messages. \\*
& 2.4.2 Commit Granularity \& Strategy & 54
& Keep each logical change in a separate commit. \\*
& 2.4.3 Pre-commit checks & 127
& Run lint and tests before committing. \\
\midrule

\rowcolor{headerbg}\multicolumn{4}{c}{\textbf{C3 Modification Scope \& Change Budget}} \\
\midrule
3.1 Spatial scope
& 3.1.1 Directory-level constraints & 77
& Restrict changes to \texttt{src/parser/} only. \\*
& 3.1.2 File-level constraints & 28
& Modify only \texttt{config.ts} and \texttt{parser.ts}. \\*
& 3.1.3 Code-level constraints & 2
& Change only the function body, not the function signature. \\
\midrule
3.2 Quantity budget
& 3.2.1 File-count limits & 42
& Modify at most three files. \\*
& 3.2.2 Line-count limits & 84
& Keep the total diff under 50 lines. \\
\midrule
3.3 Semantic scope
& 3.3.1 Task Boundary & 158
& Fix this bug only; do not introduce refactoring or new features. \\*
& 3.3.2 Interface compatibility & 127
& Preserve the existing HTTP API path and response format. \\
\midrule

\rowcolor{headerbg}\multicolumn{4}{c}{\textbf{C4 Code Style \& Engineering Standards}} \\
\midrule
4.1 Formatting
& 4.1.1 Lint \& Formatter Tools & 30
& Ensure the new code passes \texttt{ruff format} and \texttt{ruff check}. \\*
& 4.1.2 Whitespace \& Layout Rules & 3
& Use two-space indentation rather than tabs. \\*
& 4.1.3 Syntax Format Preferences & 3
& Use single quotes consistently. \\
\midrule
4.2 Naming
& 4.2.1 Naming style & 10
& Use \texttt{snake\_case} for Python variables. \\*
& 4.2.2 Naming quality & 12
& Avoid non-descriptive variable names such as \texttt{tmp} or \texttt{data}. \\
\midrule
4.3 Type system
& 4.3.1 Type annotation requirements & 9
& Add explicit return type annotations for all new functions. \\*
& 4.3.2 Type-safety restrictions & 27
& Do not use \texttt{Any} or \texttt{as any}. \\
\midrule
4.4 Comments \& documentation
& 4.4.1 Inline comments & 13
& Add comments for complex logic, explaining why rather than what. \\*
& 4.4.2 Docstrings & 8
& Add a Google-style docstring for each new public function. \\*
& 4.4.3 Project documentation & 26
& Update the README when adding a new public API. \\
\midrule
4.5 Code structure
& 4.5.1 Modularity & 73
& Extract duplicated logic into a shared helper function. \\*
& 4.5.2 Complexity limits & 1
& Keep nesting depth below three levels. \\*
& 4.5.3 Design patterns & 8
& Prefer composition over inheritance. \\
\midrule

\rowcolor{headerbg}\multicolumn{4}{c}{\textbf{C5 Code Implementation}} \\
\midrule
5.1 Algorithm \& strategy
& 5.1.1 Algorithm Selection & 37
& Use binary search rather than linear search. \\*
& 5.1.2 Concurrency \& Synchronization Strategy & 2
& Use \texttt{asyncio} rather than threads for this task. \\*
& 5.1.3 Retry \& Fault Tolerance Strategy & 2
& Apply exponential backoff, with at most three retries. \\*
& 5.1.4 Error handling & 115
& Do not use bare \texttt{except}; catch specific exceptions. \\*
& 5.1.5 Side-effect control & 32
& Do not mutate the input object; return a new object instead. \\
\midrule
5.2 Coding style
& 5.2.1 Syntax restrictions & 3
& Do not use the walrus operator. \\*
& 5.2.2 Paradigm requirements & 1
& Use \texttt{async/await} instead of callbacks. \\*
& 5.2.3 Language features & 41
& Use f-strings rather than \texttt{.format()}. \\
\midrule
5.3 Performance requirements
& 5.3.1 Complexity constraints & 5
& Keep time complexity within \(O(n \log n)\). \\*
& 5.3.2 Resource constraints & 10
& Keep memory usage below 256 MB per request. \\*
& 5.3.3 Data processing strategy & 12
& Stream large files instead of loading them fully into memory. \\
\midrule
5.4 Technology choice
& 5.4.1 Required libraries & 12
& Use \texttt{httpx} rather than \texttt{requests}. \\*
& 5.4.2 Forbidden libraries & 11
& Do not introduce \texttt{pandas} for this simple transformation. \\
\midrule

\rowcolor{headerbg}\multicolumn{4}{c}{\textbf{C6 Environment}} \\
\midrule
6.1 Runtime versions
& 6.1.1 Language Version & 18
& Keep the implementation compatible with Python 3.10 and 3.11. \\*
& 6.1.2 Framework \& Platform Version & 4
& Maintain compatibility with PostgreSQL 14+. \\
\midrule
6.2 Platform compatibility
& 6.2.1 Path \& Filesystem & 3
& Do not hard-code \texttt{/tmp}; use a portable temporary directory API. \\*
& 6.2.2 Encoding \& Runtime Environment & 5
& Assume UTF-8 explicitly rather than relying on the system default encoding. \\
\midrule
6.3 Isolation
& 6.3.1 Filesystem isolation & 28
& Do not write files outside the repository workspace. \\*
& 6.3.2 System environment isolation & 12
& Do not modify global \texttt{git} or \texttt{npm} configuration. \\
\midrule

\rowcolor{headerbg}\multicolumn{4}{c}{\textbf{C7 Dependency \& Build Strategy}} \\
\midrule
7.1 Dependency changes
& 7.1.1 Addition Restrictions & 41
& Do not add any new third-party dependency. \\
\midrule
7.2 Version management
& 7.2.1 Lock File Policy & 4
& Do not modify \texttt{poetry.lock}. \\*
& 7.2.2 Version Range Policy & 1
& Allow patch upgrades but not major-version upgrades. \\
\midrule
7.3 Build pipeline
& 7.3.1 CI/CD configuration & 1
& Do not modify files under \texttt{.github/workflows/}. \\*
& 7.3.2 Build Scripts \& Output & 2
& Do not change the build artifact path or filename. \\
\midrule

\rowcolor{headerbg}\multicolumn{4}{c}{\textbf{C8 Security \& Dangerous Operations}} \\
\midrule
8.1 Dangerous commands \& privileges
& 8.1.1 Destructive commands & 24
& Do not use \texttt{rm -rf} or \texttt{git reset --hard}. \\*
& 8.1.2 Privilege \& System Operations & 14
& Do not use \texttt{sudo} or modify firewall rules. \\
\midrule
8.2 Information leakage
& 8.2.1 Secret \& Credential Protection & 3
& Do not print API keys or tokens in logs. \\*
& 8.2.3 System Information Protection & 2
& Do not expose internal server paths in error output. \\
\midrule
8.3 Code security
& 8.3.1 Injection protection & 5
& Use parameterised SQL rather than string concatenation. \\*
& 8.3.3 Encryption \& Secure Coding & 1
& Use \texttt{bcrypt} or \texttt{argon2} for password hashing. \\
\midrule

\rowcolor{headerbg}\multicolumn{4}{c}{\textbf{C9 Testing}} \\
\midrule
9.1 Test writing
& 9.1.1 Test Coverage Requirements & 156
& Add a regression test that reproduces the bug. \\*
& 9.1.2 Test quality standards & 104
& Ensure each test is independent and order-insensitive. \\
\midrule
9.2 Test execution
& 9.2.1 Test Suite Selection & 155
& Run the affected unit tests before the full suite. \\*
& 9.2.2 Static Analysis \& Checks & 45
& Run \texttt{mypy} and \texttt{ruff check} before submission. \\
\midrule
9.3 Test reporting
& 9.3.1 Result summary & 4
& Report which tests passed or failed, rather than saying only ``tests passed.'' \\*
& 9.3.2 Reproduction \& Comparison & 66
& Show the before/after test results for the fix. \\
\midrule

\rowcolor{headerbg}\multicolumn{4}{c}{\textbf{C10 Output Format \& Reporting}} \\
\midrule
10.1 Language \& tone
& 10.1.1 Language Choice & 62
& Answer in Chinese throughout. \\*
& 10.1.2 Tone \& Style & 66
& Avoid emojis and keep the response concise. \\
\midrule
10.2 Structure \& templates
& 10.2.1 Report Template & 5
& Organise the final report as: reproduce / root cause / fix / verification / risk / rollback. \\
\midrule
10.3 Citation \& evidence
& 10.3.1 Code References & 45
& Cite the modified file path and line numbers. \\*
& 10.3.2 Execution Evidence & 53
& Include the relevant command output as evidence for each conclusion. \\

\end{longtable}
}

\section{Additional Methodological Details}
\label{app:construction_details}

\subsection{Taxonomy Design Principles}

The ANCHOR taxonomy was not designed purely from an a priori schema. Instead,
we started from grounded open-label annotation over real developer interactions
with coding agents, including task instructions, inline review comments, and
post-hoc feedback on agent trajectories. Senior software engineers then
iteratively clustered near-duplicate labels, resolved boundary cases, and
stress-tested the emerging hierarchy against held-out trajectories. The goal of
this refinement process was not merely descriptive coverage, but an operational
taxonomy that could support scalable instantiation, evaluation, and error
analysis.

Five design principles guided this process. First, \emph{coverage}: the
taxonomy should cover the major engineering constraint types that appear in
practice, so that real developer instructions map into it with minimal residue.
Second, \emph{semantic exclusivity}: leaf categories should be as disjoint as
possible, reducing ambiguity during annotation and automatic mapping. Third,
\emph{operationalisability}: each leaf should admit concrete, testable
instantiations with stable pass conditions under either rule-based or
LLM-based evaluation. Fourth, \emph{hierarchical organisation}: the taxonomy
should support both fine-grained diagnostic analysis at the leaf level and
coarse-grained reporting at the category level. Fifth, \emph{language
neutrality}: the same conceptual categories should transfer across the
multilingual repository settings covered by ANCHOR.

The current release instantiates this design as a three-level hierarchy with
10 top-level categories, 34 mid-level subcategories, and 78 leaf categories.
At a coarse level, these groups capture complementary dimensions of engineering
behaviour: process- and trajectory-level constraints (C1--C2), patch- and
implementation-level constraints (C3--C5), environment and system-boundary
constraints (C6--C8), and validation and reporting requirements (C9--C10).
Leaf categories are the smallest operational unit: each is intended to capture
a semantically coherent developer requirement that can be instantiated into
task-grounded constraints and evaluated with relatively stable verification
logic.

\subsection{Environment Curation and Context Compression}

The construction pipeline begins by selecting repository tasks that can support
realistic and verifiable engineering constraints. The filtering objective is
not to privilege easier tasks, but to ensure that each retained instance admits
trajectory-grounded evaluation. In practice, we prioritise instances with
(i) a usable execution environment, (ii) sufficient task and repository context
to support downstream constraint selection, (iii) moderate patch complexity
rather than trivial micro-edits or very long-horizon changes, and (iv) broad
coverage across programming languages and repository types.

For each retained instance, we collect the repository identifier, target
language, original problem statement, gold patch, and repository-level
documentation from DeepWiki~\citep{deepwiki2025}. These artefacts provide
complementary signals for later stages: the problem statement and patch reveal
the task objective and likely affected code regions, while repository
documentation provides project-level context such as architecture, tooling,
testing practices, and engineering conventions.

Raw repository documentation is often too long and noisy to include directly in
downstream prompts. We therefore compress it into a compact project-context
card that preserves information useful for constraint construction, such as
repository purpose, technology stack, directory structure, testing practices,
and version information, while discarding low-value content such as diagrams,
verbose API documentation, illustrative code examples, marketing language, and
redundant passages. This step is not merely length reduction: it is a targeted
information-distillation stage that gives later prompts repository-level
evidence without overwhelming the context budget.

\subsection{Subset Selection and Atomic Instantiation}

Given the full taxonomy, the next challenge is deciding which leaf categories
are relevant to a particular task. ANCHOR treats this as a high-recall
selection problem. The coarse selection stage receives the full taxonomy, the
repository identifier, programming language, problem statement, compressed
repository context, and the gold patch. The prompt permits the gold patch to inform task context and category
selection, while explicitly forbidding requirements that copy its exact
counts or restate the reference solution. This is a prompt-level safeguard,
not evidence that category selection is independent of the patch.

Selection decisions are based on three complementary evidence sources. The
first is \emph{task-type evidence}: some categories are more naturally
associated with certain task types, such as testing or interface-compatibility
constraints in bug-fixing tasks. The second is \emph{repository-context
evidence}: formatter, linter, type-checking, or dependency-management
conventions make some categories more meaningful for a given project. The third
is \emph{modification-scope evidence}: tasks touching public interfaces,
filesystem logic, exception handling, or security-sensitive code naturally
raise different constraint families. At the same time, the selector avoids
retaining categories that are only theoretically applicable to almost any
coding task, because these tend to yield generic and weakly discriminative
constraints later on.

This stage is intentionally followed by an implicit precision filter during
atomic instantiation. A selected category is retained in the final benchmark
instance only if it can yield one or more constraints that are naturalistic,
non-blocking, specific and verifiable, task-grounded, and independent of
the reference implementation in content. Categories that would only produce generic, overly restrictive,
or weakly diagnostic constraints are dropped. In this sense, ANCHOR uses a
recall--precision cascade: coarse selection aims to avoid prematurely missing
meaningful categories, while atomic instantiation enforces quality control over
what ultimately appears in the benchmark.

\subsection{Composite Constraints and Query Synthesis}

Real developer instructions are often not flat lists of independent
requirements. They may specify joint satisfaction, required execution order, or
precedence among competing requirements. ANCHOR captures this structure by
organising selected subsets of atomic constraints into connected composite
graphs. Edges carry one of three typed relations: \texttt{AND} requires joint
satisfaction, \texttt{THEN} requires both satisfaction and correct temporal
order, and \texttt{PRIORITY} requires the source constraint to take precedence
over the target when the two are in tension.

Composite constraints are introduced only when the underlying relations are
semantically natural and task-meaningful; ANCHOR does not force unrelated
constraints into a shared graph merely to increase structural complexity. Not
all atomic constraints must participate in composites, and standalone
constraints remain as independent instruction items. Edge types may be mixed
within one composite, and each atomic constraint participates in at most one
composite, so composites partition the related atomics into disjoint,
self-contained instruction units. This connected-subgraph
formulation keeps each composite interpretable as a single instruction unit
rather than an arbitrary bundle of checklist clauses.

The final query-synthesis stage converts the structured representation into a
natural developer instruction suitable for direct agent execution. Each query is
written in two parts: a short task description rewritten in the voice of a
developer assigning work, and a constraint list composed of separable
instruction items. Independent atomic constraints become individual items,
while each composite constraint is rendered as a single natural-language item
that preserves the relation semantics among its member atomics. A key
requirement of this stage is specificity preservation: exact file paths,
command names, flags, module references, and behavioural conditions from the
instantiated constraints must remain intact. The synthesis step must also avoid
introducing new constraints or leaking gold-solution details. Prompt templates
for these stages are provided in
Appendix~\ref{app:prompts:compress}--\ref{app:prompts:query}.

\section{Prompt Templates}

\subsection{Context Compression Prompt}
\label{app:prompts:compress}
The released prompt directory does not contain the repository-context
compression template. In the current pipeline, this step uses a shorter
summarization instruction to distill raw DeepWiki documentation into a compact
project-context card covering repository purpose, technology stack, directory
structure, engineering conventions, testing practices, and version
compatibility. The prompt templates below are included verbatim from the
released prompt set, with variable slots preserved in braces.

\subsection{Coarse Constraint Selection Prompt}
\label{app:prompts:selection}
\tcbinputlisting{promptstyle,title={\small\bfseries Constraint subset selection prompt},listing file={prompts/subset_selection_prompt.txt}}

\subsection{Atomic Constraint Instantiation Prompt}
\label{app:prompts:atomic}
\tcbinputlisting{promptstyle,title={\small\bfseries Atomic constraint instantiation and composite composition prompt},listing file={prompts/atomic_instantiation_prompt.txt}}

\subsection{Query Synthesis Prompt}
\label{app:prompts:query}
\tcbinputlisting{promptstyle,title={\small\bfseries Query synthesis prompt},listing file={prompts/query_generation_prompt.txt}}

\subsection{Checklist Generation Prompt}
\label{app:prompts:checklist}
\tcbinputlisting{promptstyle,title={\small\bfseries Checklist generation prompt},listing file={prompts/checklist_generation_prompt.txt}}

\subsection{LLM Judge Prompt}
\label{app:prompts:judge}
\tcbinputlisting{promptstyle,title={\small\bfseries LLM judge prompt},listing file={prompts/llm_judge_prompt.txt}}

\subsection{Error Attribution Prompt}
\label{app:prompts:error}
\tcbinputlisting{promptstyle,title={\small\bfseries Error attribution prompt},listing file={prompts/error_attribution_prompt.txt}}

\end{document}